\documentclass{article} % For LaTeX2e
\usepackage[preprint]{colm2025_conference}

\usepackage{microtype}
\usepackage{hyperref}
\usepackage{url}
\usepackage{booktabs}

\usepackage{lineno}

\usepackage{amsfonts}       % blackboard math symbols
\usepackage{nicefrac}       % compact symbols for 1/2, etc.
\usepackage{microtype}      % microtypography
\usepackage{xcolor}         % colors

\usepackage[toc,page,header]{appendix}
\usepackage{minitoc}

\usepackage{titletoc}
\usepackage{natbib}
\usepackage{graphicx}
\usepackage{booktabs}
\usepackage{enumitem}
\usepackage{tabularray}
\usepackage{amsmath}
\usepackage{pifont}
\usepackage{wrapfig}
\usepackage{caption,graphicx,newfloat}
\DeclareCaptionType{Prompt}
\DeclareCaptionType{Example}

\usepackage{amsmath}
\usepackage{amssymb}
\usepackage{mathtools}
\usepackage{amsthm}

\newcommand{\update}[1]{\textcolor{black}{#1}}

\usepackage[capitalize,noabbrev]{cleveref}

\theoremstyle{plain}
\newtheorem{theorem}{Theorem}[section]

\newtheorem{lemma}[theorem]{Lemma}

\theoremstyle{definition}
\newtheorem{definition}[theorem]{Definition}

\theoremstyle{remark}

\definecolor{darkblue}{rgb}{0, 0, 0.5}
\hypersetup{colorlinks=true, citecolor=darkblue, linkcolor=darkblue, urlcolor=darkblue}

\title{Data-Centric Human Preference with Rationales for Direct Preference Alignment  }

\author{%
Hoang Anh Just\\
Virginia Tech\\
\texttt{just@vt.edu}\\
\And
Ming Jin\\
Virginia Tech\\
\texttt{jinming@vt.edu}\\
\And
Anit Sahu\\
Amazon\\
\texttt{anit.sahu@gmail.com}\\
\And
Huy Phan\\
Amazon\\
\texttt{huypq@amazon.co.uk}\\
\And
Ruoxi Jia\\
Virginia Tech\\
\texttt{ruoxijia@vt.edu}\\
}

\begin{document}

\ifcolmsubmission
\linenumbers
\fi

\maketitle

\begin{abstract}

Aligning language models with human preferences through  reinforcement learning from human feedback is crucial for their safe and effective deployment. The human preference is typically represented through comparison where one response is chosen over another for a given prompt. However, standard preference datasets often lack explicit information on why a particular choice was made, presenting an ambiguity that can hinder efficient learning and robust alignment, especially given the high cost of acquiring extensive human annotations. While many studies focus on algorithmic improvements, this work adopts a data-centric perspective, exploring how to enhance learning from existing preference data. We propose augmenting standard preference pairs with rationales that explain the reasoning behind the human preference. Specifically, we introduce a simple and principled framework that leverages machine-generated rationales to enrich preference data for preference optimization algorithms. Our comprehensive analysis demonstrates that incorporating rationales improves learning efficiency. Extensive experiments reveal some advantages: rationale-augmented learning accelerates convergence and can achieve higher final model performance. Furthermore, this approach is versatile and compatible with various direct preference optimization algorithms. Our findings showcase the potential of thoughtful data design in preference learning, demonstrating that enriching existing datasets with explanatory rationales can help unlock improvements in model alignment and annotation efficiency.

% Reinforcement learning from human feedback plays a crucial role in aligning language models towards human preferences, traditionally represented through comparisons between pairs or sets of responses within a given context. While many studies have enhanced algorithmic techniques to optimize learning from such data, this work shifts focus to improving preference learning through a data-centric approach. Specifically, we propose enriching existing preference datasets with machine-generated rationales that explain the reasons behind choices. We develop a simple and principled framework to augment current preference learning methods with \emph{rationale} information. Our comprehensive analysis highlights how rationales enhance learning efficiency. Extensive experiments reveal that rationale-enriched preference learning offers multiple advantages: it improves annotation efficiency, accelerates convergence to higher-performing models, and reduces verbosity bias and hallucination. Furthermore, this framework is versatile enough to integrate with various preference optimization algorithms. Overall, our findings highlight the potential of re-imagining data design for preference learning, demonstrating that even freely available machine-generated rationales can significantly boost performance across multiple dimensions

\end{abstract}

\section{Introduction} \label{sec: intro}

Human preference optimization is a critical stage in preparing large language models (LLMs) for real-world deployment. Its primary goal is to align model behavior with human expectations and prevent undesirable outputs~\citep{christiano2017deep, stiennon2020learning, bakker2022fine}. This alignment process typically relies on datasets containing prompts paired with ranked responses, which encapsulate human preferences. Popular methods like Reinforcement Learning from Human Feedback (RLHF)\citep{ouyang2022training} leverage this data by first training a reward model to mimic human judgments and then using reinforcement learning\citep{schulman2017proximal} to optimize the LLM policy against this reward signal. More recently, direct preference optimization (DPO)~\citep{rafailov2024direct} and its variants offer an alternative approach, directly optimizing the policy using preference pairs through a loss function derived from an implicit reward model, thus avoiding the complexities of training an explicit reward model.

Despite their effectiveness, these methods face significant challenges. Models can overfit to the preference dataset~\citep{azar2024general}, suffer performance degradation on broader tasks~\citep{pal2024smaug}, learn to exploit the rewards~\citep{amodei2016concrete}, or generate excessively long and unhelpful responses~\citep{park2024disentangling}. Furthermore, acquiring high-quality human preference data is notoriously costly and time-consuming~\cite{zheng2023judging, tan2024large, liu2024direct, zhou2025self}. Yet, insufficient data can lead to underfitting, failing to achieve robust alignment~\cite{jinnai2024annotation}. Much research has focused on mitigating these issues through algorithmic advancements, such as regularizing the optimization objective~\citet{pal2024smaug, amini2024direct, park2024disentangling} or developing entirely new learning formulations~\citet{ethayarajh2024kto, hong2024reference, yuan2023rrhf, munos2023nash, swamy2024minimaximalist, wu2024self}. While recent techniques such as SimPO~\citep{meng2024simpo} and SPPO~\citep{wu2024self} have demonstrated remarkable performance gains, often enabling smaller models to outperform larger ones on benchmarks like AlpacaEval~\cite{dubois2024length}, these improvements largely stem from modifications to the objective function or the generation of entirely new preference pairs.

% SimPO~\citep{meng2024simpo} and SPPO~\citep{wu2024self} have demonstrated remarkable results on small models achieving high AlpacaEval benchmark scores often outperformaing larger models. However, these improvements came either from algorithmic improvements by modyfying the objective function or by completely generating new preference pair datasets. 
% However, as human preference data are valuable and our objective is to align with the human data, in our study, we would like to focus on further improving the model alignment on human preference data with direct preference optimization methods via a data-centric method, in contrast to the algorithmic improvements. How can we help the model better learn from existing human preference data to align with human expectation?

However, human preference data are inherently valuable, representing the ground truth for alignment, and acquiring more is often prohibitively expensive. Generating synthetic data, while scalable, risks introducing biases or deviating from nuanced human intentions. This motivates a shift in focus: instead of solely pursuing algorithmic refinements or creating new datasets, can we extract more value from the existing, carefully curated human preference data? Our study adopts this data-centric perspective, specifically exploring how to enhance model alignment using direct preference alignment methods by improving the way models learn from the available human preference data. We pose the central question: How can we help the model better learn from existing human preference data to align more effectively with human expectations?

% In our study, we transition from an algorithmic to a data-centric approach, posing the key question: \emph{How can enhancing the preference dataset aid the model in boosting its performance and efficiency in preference learning?} By rethinking preference learning through the lens of data, we aim to discover new insights and opportunities that can unlock the potential for more robust, data-efficient preference learning.

% However, our goal is to learn from the provided human preference data, when we generate new preference pairs, the expectation might be different from original. One possible way to improve human alignment is to further acquire more high-quality human preference data. However, that comes at a high cost often crossing 10,000 dollars, and might not be available at hand. Therefore, our goal is to learn more from already provided preference data and to help the model better align with it.

% address data-centric questions. Are the preference data adequately represented and efficiently used? How can we enrich the current data setup to assist the model in improving its performance and efficiency in preference learning? By re-examining the aspect of data, we strive to uncover new insights and opportunities for enhancing preference learning models, potentially leading to more accurate and efficient solutions.
% Adding more control as to what behavior we expect from the model to follow with. 

A key limitation of current preference datasets lies in their opacity regarding the reasons behind a preference. Given a prompt and two responses, why is one preferred over the other? While the distinction may be obvious in some cases (e.g., factual correctness), it becomes highly ambiguous when responses are closely matched in quality or exhibit subtle trade-offs. Superficial features like response length offer unreliable cues; longer responses might be favored for comprehensiveness in one context, while shorter ones are preferred for conciseness in another. Furthermore, human preferences can be subjective and stem from diverse, unstated reasoning. Without explicit explanations explaining these underlying criteria, the model faces a significant learning challenge. This ambiguity can lead to data inefficiency, requiring more examples to discern patterns, or worse, cause the model to learn spurious correlations, hindering true alignment and potentially degrading performance.

% Given the current setup of preference datasets, an important question arises: why would a certain response be preferred over another? For obvious cases, it is simple for humans to understand the preference. However, when the responses are closely matched, understanding the preference without any explanation becomes challenging. Even superficial features such as length might not serve as a straightforward metric to determine preference. In one instance, a longer response may be favored for its comprehensiveness, while in another instance, a shorter response might be preferred for its conciseness. Another consideration regarding preferences is that individuals might have varying preferences for different reasons, and without explicitly outlining these reasons, one would be unable to discern the underlying rationale. Given these challenges, the model will struggle to learn these preferences without any explanations, causing data inefficiency, and worse, it could learn the wrong cues, decreasing performance.

% \begin{wrapfigure}{r}{0.45\textwidth}
\begin{figure}[h]
  \begin{center}
    \includegraphics[width=0.70\linewidth]{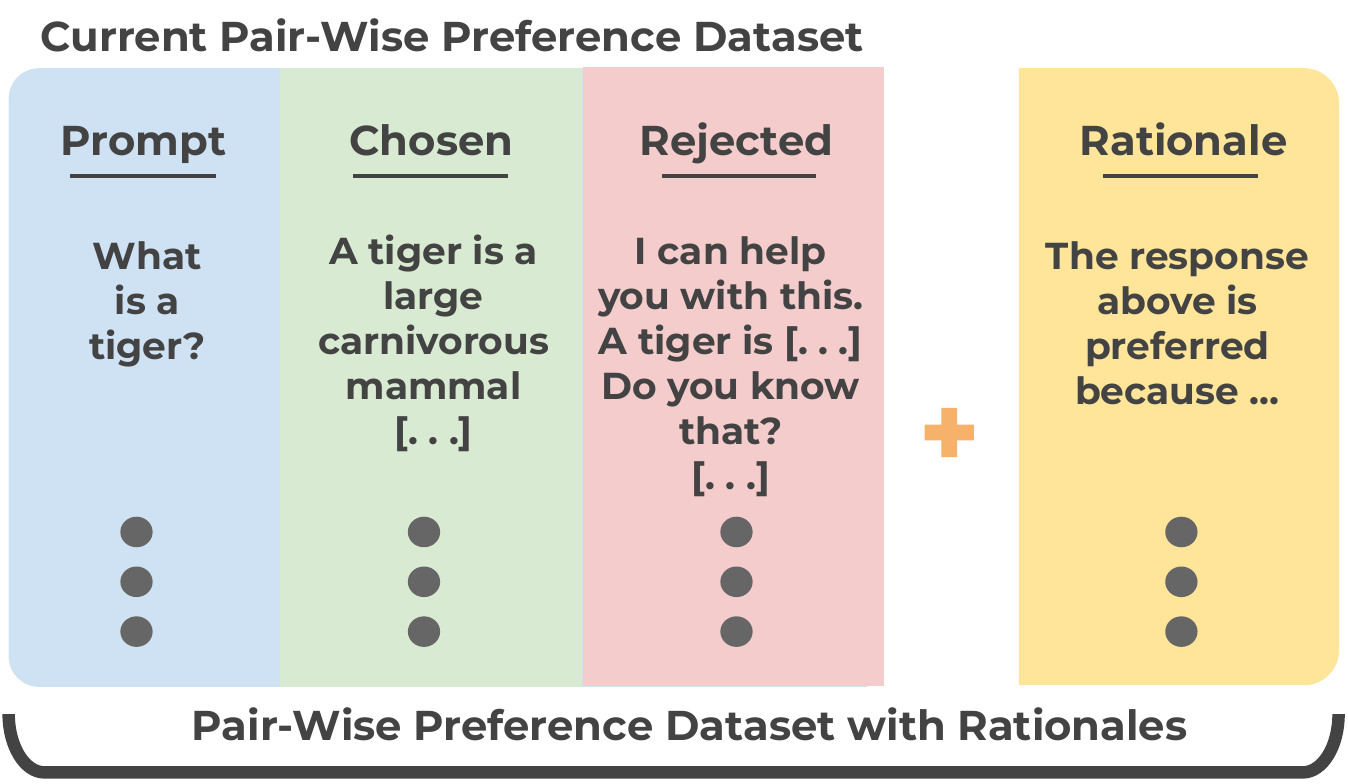}
  \end{center}
  \vspace{-1em}
  \caption{Comparison between the current pair-wise preference dataset used for preference learning and the enriched dataset with added  rationales.}
  % \ruoxi{shall we move it to intro}}
    \label{fig:dataset with rationale}
\end{figure}
% \end{wrapfigure}

For the reasons outlined above, we propose a natural extension to the standard preference dataset structure by incorporating rationales. We define a rationale as an explanation accompanying each preference pair, explaining why the chosen response was preferred over the rejected one for the given prompt. This approach aims to provide the model with deeper contextual understanding during preference learning. This idea also draws inspiration from social studies~\citep{mitchell1986explanation, chi1994eliciting,crowley1999explanation, williams2010role}, showing that adding explanations to answers improves one's understanding of the problem compared to individuals who do not provide explanations. 

% For the reasons outlined above, we propose a natural extension to the current dataset structure by enriching the preferences with rationales, aiding the model in better understanding during preference learning. Rationales explain why one response is preferred over another for a given prompt. This idea also draws inspiration from social studies~\citep{mitchell1986explanation, chi1994eliciting,crowley1999explanation, williams2010role}, showing that adding explanations to answers improves one's understanding of the problem compared to individuals who do not provide explanations. 

We hypothesize that by enriching existing human preference datasets with rationales, models undergoing direct preference alignment can develop a more nuanced understanding of the underlying human values. This enhanced comprehension is expected to translate into improved preference modeling, higher performance on alignment benchmarks, and greater annotation efficiency, allowing models to learn human preferences more effectively from the available annotated data before potentially needing to resort to further costly data collection.

\textbf{Contributions.} This paper provides a new data-centric perspective on direct preference learning. We list the summary of contributions:
% \vspace{-0.5em}
\begin{itemize}[leftmargin=*]
    \setlength{\parskip}{0pt}
    \setlength\itemsep{0em}
    \item[\ding{121}] We introduce rationales into the human preference learning framework, where rationales explain the reasons behind the preference for a particular response and derive a straightforward formulation for the preference function to extend the rationales and show how to adapt our method to current direct preference learning algorithms such as DPO.
    % In practice, these rationales can be generated in a cost-effective manner by prompting an off-the-shelf language model, which may or may not have undergone preference learning.
    % \item[\ding{121}] We derive a straightforward formulation for the preference function to extend the rationales and show how to adapt our method to current direct preference learning algorithms such as DPO.
    \item[\ding{121}] We analytically examine the impact of the rationale on preference training through the lens of information theory and empirically show the impact of preference learning with rationales.
\end{itemize}
% \vspace{-0.5em}
In a broader context, our approach presents a new paradigm for data-centric research in language modeling: rather than focusing on pruning samples to distill the most informative pieces from a dataset~\cite{albalak2024survey}, we explore how to enrich each sample's information content and examine its impact. The promising results presented in this paper demonstrate the effectiveness of enhancing individual samples' information content in preference learning and suggest that this approach may hold potential for improving learning in other domains.

\section{Related Work (extended version in Appendix~\ref{app: ext rel work})} \label{sec: rel work}

% Method
% Problems
% Solutions

Aligning large language models (LLMs) using techniques derived from human preferences is an important step for current LLMs~\citep{casper2023open, christiano2017deep, ziegler2019fine, stiennon2020learning}. While initial methods often relied on explicit reward modeling followed by reinforcement learning algorithms like Proximal Policy Optimization (PPO)\citep{schulman2017proximal, ouyang2022training}, newer approaches such as Direct Preference Optimization (DPO)\citep{rafailov2024direct} bypass explicit reward models. Despite their success, both PPO-based RLHF and DPO face challenges, such as overfitting to the preference data, underfitting due to data scarcity~\citep{azar2024general, pal2024smaug}, and reward hacking~\citep{amodei2016concrete}.
% Much subsequent research has focused on algorithmic enhancements to address these limitations. 
To address these issues, researches introduce length-regularized DPO (R-DPO) \citep{park2024disentangling}, odds ratio preference optimization (ORPO) \citep{hong2024reference}, Kahneman-Tversky optimization (KTO) \citep{ethayarajh2024kto}, or ranked responses methods \citep{yuan2023rrhf}. General preference modeling, moving beyond Bradley-Terry assumptions \citep{bradley1952rank, bertrand2023limitations}, offers alternatives such as Nash preference optimization \citep{munos2023nash, swamy2024minimaximalist, rosset2024direct, wu2024self}. However, these methods focus primarily on algorithmic enhancements. A separate line of data-centric work aims to improve alignment by synthesizing entirely new preference datasets, often leveraging stronger models or self-critique mechanisms~\citep{yangrlcd, pace2024west, meng2024simpo, wu2024self, wang2024self}.
Our work pursues a distinct, orthogonal approach. Instead of modifying the optimization algorithm or generating new preference pairs, we focus on enhancing the utility of existing preference datasets for direct optimization methods by introducing rationales that explain why one response was preferred. While related data-centric concepts which incorporate critiques to preference data have shown promise, particularly to refine reward models~\citep{ye2024improving, ankner2024critique, yu2024self}, our rationales are designed for direct preference optimization methods and differ from critiques in design. Particularly, our rationales explain the difference between the responses, the critiques are made for each preference individually and independent of other preference. We provide an extended discussion of related work in the Appendix~\ref{app: ext rel work}.

\section{Method} \label{sec: method}

In this section, we introduce the incorporation of rationales into preference learning and show the derivation of adapting current methods. We present a demonstration of extending the direct preference optimization (DPO) algorithm to incorporate the rationales, while similar extensions can be applied to other variants of DPO. 
% \ruoxi{emphasize we focus on DPO as a pilot study and similar derivations can be done to extend the rationale based on learning to other variants of DPO} to incorporate rationales into training.  
Further, we analyze theoretically the possible impact of rationales through the perspective of information theory.

% \begin{figure*}
% % \vspace{-2.5em}
%     \centering%
% 	\includegraphics[width=0.55\linewidth]{fig/new dataset rationale.pdf}
% 	\caption{Comparison of current pair-wise preference dataset used for preference learning with the augmented dataset with added rationales.}
%     \label{fig:dataset with rationale}
% \end{figure*}

\subsection{Preliminaries} 
% \vspace{-1em}
\paragraph{Notations.} 
Let $\mathcal{D}$ denote the pair-wise preference dataset of size $N$, $\mathcal{D} = \{ x^{(i)}, y_w^{(i)}, y_l^{(i)}\}_{i=1}^N $, where $x^{(i)}$ is a context, $y_w^{(i)}$ is the preferred/chosen/winning response to the context $x^{(i)}$ over the unpreferred/rejected/losing response $y_l^{(i)}$. Let $\pi_{\theta}$ and $\pi_{\text{ref}}$ denote the policy to be preference optimized and the reference policy respectively. In our setting, the policies are the language model to be preference trained and the base or supervised fine-tuned SFT model, respectively. To compute the joint probability of the autoregressive language model $\pi$ generating the response $y$ given the prompt $x$, we compute the product of probabilities after observing each token: $\pi(y|x) = \Pi_{t=0}^{|y|}  \pi (y_t | x, y_{0:t})$.

\paragraph{Reward Modeling with DPO.} \label{sec:dpo}
In the RLHF process~\citep{christiano2017deep, ziegler2019fine,stiennon2020learning,bai2022training,ouyang2022training,rafailov2024direct}, the goal is to align the language model towards human preferences. The preferences ranking from the dataset $\mathcal{D}$ is assumed to be sampled from the latent reward function $r(x,y)$ and the preference function is assumed to be generated by the Bradley-Terry model~\citep{bradley1952rank}: $p^*(y_w \succ y_l | x) = \sigma (r^*(x,y_w) - r^*(x,y_l))$, where $\sigma$ is the sigmoid function. The reward function then can be estimated by minimizing the log-likelihood of the following objective
% , $\log p (y_w \succ y_l | x)$, which can be written as follows 
$\mathcal{L} (r, \mathcal{D}) = -\mathbb{E}_{(x,y_w,y_l) \sim \mathcal{D}} \left[ \log \sigma(r(x,y_w) - r(x,y_l)) \right]$. Then the next step is to tune the language model with the reward model as follows by maximizing the rewards and not diverging from the fixed reference model: $\max_{\pi_{\theta}} \mathbb{E}_{x\sim \mathcal{D}, y \sim \pi_{\theta}(y | x)} \left[ r(x,y)\right] - \beta \mathbb{D}_{\text{KL}} \left[ \pi_{\theta} (y |x) \| \pi_{\text{ref}}(y|x) \right]$, where $\beta$ is a hyperparameter measuring the divergence between two policies. 
Alternatively, with a reparametrization of the Bradley-Terry preference model~\citep{rafailov2024direct}, the preference function can be expressed in terms of policy $\pi^*$:
{\tiny
\begin{align} \label{eq: pairwise}
    p^*(y_w \succ y_l | x) = \sigma \left( \beta \log \frac{\pi^*(y_w |x)}{\pi_{\text{ref}}(y_w|x)} - \beta \log \frac{\pi^*(y_l |x)}{\pi_{\text{ref}}(y_l|x)} \right).
\end{align}}
Thus, to estimate the policy, \citet{rafailov2024direct} proposes to directly minimize the log-likelihood of the following DPO loss:
{\tiny$\mathcal{L}(\pi_{\theta};\pi_{\text{ref}}) = -\mathbb{E}_{(x,y_w,y_l)\sim \mathcal{D}} \left[ \log \sigma \left( \beta \log \frac{\pi_{\theta}(y_w |x)}{\pi_{\text{ref}}(y_w|x)} - \beta \log \frac{\pi_{\theta}(y_l |x)}{\pi_{\text{ref}}(y_l|x)} \right) \right] $}.

% proposed a direct one-step optimization of the language model with implicit reward modeling, where $\pi^*$ is the optimal policy:

% Then the DPO loss becomes: $\mathcal{L}(\pi_{\theta};\pi_{\text{ref}}) = -\mathbb{E}_{(x,y_w,y_l)\sim \mathcal{D}} \left[ \log \sigma \left( \beta \log \frac{\pi_{\theta}(y_w |x)}{\pi_{\text{ref}}(y_w|x)} - \beta \log \frac{\pi_{\theta}(y_l |x)}{\pi_{\text{ref}}(y_l|x)} \right) \right] $.
% the reward modelling
% Given the preference dataset $\mathcal{D}$, the ranking of responses are assumed 

% The DPO has developed an efficient way to optimize the policy to optimize the reward modeling with the Bradley-Terry preference model assumption as follows:

% di
% show dpo formulation

\subsection{Formulation of Preference Learning with Rationales} \label{sec: rationale formulation}

% general formulation for preference learning with rationales

% We introduce a 

% The preference dataset 

Although preferences are modeled given the preferred and unpreferred responses, there are nuances in the responses that are obscure for the model to comprehend and catch the differences between them. Therefore, our goal is to help the model learn preferences by providing guidance cues in the preference tuning process, which we call \emph{rationales}. The rationales explain why a given response is preferred over the other response. For that reason, we extend the current preference learning with a data-centric technique to incorporate rationales, and we term this the \emph{rationale-enriched preference function}, where the updated preference function 
% \ruoxi{how about "rationale-augmented preference function"}
is formulated as $p^*(y_w \succ y_l, r | x)$, and $r$ is the rationale from the updated dataset $\mathcal{D}' = \{ x^{(i)}, y_w^{(i)}, y_l^{(i)}, r^{(i)}\}_{i=1}^N$. By the chain rule, we arrive at:
{\tiny
\begin{align}\label{eq: rationale-enriched preference function}
    p^*(y_w \succ y_l, r | x) &=  p^*(y_w \succ y_l| x) \cdot p^*(r |x, y_w \succ y_l), 
\end{align}}
where $p^*(y_w \succ y_l| x)$ is the pair-wise preference term modeled in Section~\ref{sec:dpo}, and $p^*(r |x, y_w \succ y_l)$ is the probability of the rationale $r$ given the context $x$ and the preference $y_w \succ y_l$. Given the policy $\pi^*$, we can retrieve the probability of generating the rationale $r$ given the context $x$ and the preference $y_w \succ y_l$, $\pi^* (r | x, y_w \succ y_l)$. 
% \jin{It seems that the model you use to explain preference to generate $r$ is not the same model you aim to optimize $\pi_{\theta}$? it seems we should clarify in the following to avoid potential confusion. }\hoang{We use the same model to retrieve probabilities of generating responses as well as the rationales.}
Similarly when retrieving the probability of generating responses $y_w$ and $y_l$ for the prompt $x$, which are given in the preference dataset $\mathcal{D}'$, we can also retrieve the probability of generating rationale $r$ given $x$,$y_w$, and $y_l$, where $(x,y_w,y_l,r)\sim \mathcal{D}'$.
In practice, we ask the policy language model $\pi_\theta$ to explain why the response $y_w$ is preferred over the response $y_l$ for the prompt $x$ and retrieve the probability of generating the rationale $r$. Thus, $p(r|x,y_w \succ y_l) = \pi_{\theta}(r | x,y_w \succ y_l)$. 
% \jin{This equation is not true, since $p$ is supposed to be the true unknown distribution. We can use  $p_\theta$ instead.}\hoang{Right, I changed the notations to be at optimality.}.

% Afterward, we convert the log-probability of the sequence of rationales to the probability space, for which we use the sigmoid function in our case. However, the exploration of other functions is important to study. For generating rationales, we refer readers to Section~\ref{sec:exp}.

\paragraph{Adaptation to DPO Loss.} 

After deriving the rationale-enriched preference learning function, we extend the DPO method to incorporate rationales. By substituting Equation~\ref{eq: pairwise} and $p(r|x,y_w \succ y_l)$ into Equation~\ref{eq: rationale-enriched preference function}, we can express the rationale-enriched preference function in terms of policy.
We can then estimate the policy by optimizing $\pi_{\theta}$ through maximum likelihood using the following objective over the updated preference dataset $\mathcal{D}' = \{ x^{(i)}, y_w^{(i)}, y_l^{(i)}, r^{(i)}\}_{i=1}^N$, which we term as  rationale-DPO (RDPO):
{\tiny
\begin{align} \nonumber
    \mathcal{L}_{\text{RDPO}}(\pi_{\theta};\pi_{\text{ref}}) &= -\mathbb{E}_{(x,y_w,y_l,r)\sim \mathcal{D}'} \left[  \log \sigma \left( \beta \log \frac{\pi_{\theta}(y_w |x)}{\pi_{\text{ref}}(y_w|x)}  - \beta \log \frac{\pi_{\theta}(y_l |x)}{\pi_{\text{ref}}(y_l|x)} \right)  +  \gamma \log  \pi_{\theta} (r | x, y_w \succ y_l) \ \right],
\end{align}}
\noindent where $\gamma$ is the added hyperparameter for weighting the impact of rationales on the loss. We also provide experiments on adapting ORPO and SimPO with rationales in Section~\ref{sec:exp}.
% \jin{$\gamma\log\pi_{\theta} (r | x, y_w \succ y_l)$}
% \ruoxi{revise}

% show for the case of dpo

% derivation of our formulation
% algorithm
% hyperparemteres
% how generate rationales

% \begin{figure*}
% % \vspace{-2.5em}
%     \centering%
% 	\includegraphics[width=0.55\linewidth]{fig/markov chain modl.pdf}
% 	\caption{Markov Model with Rationale (Top) and without Rationale (Bottom), with green $S$ being the Input-Output, with yellow $R$ being rationale, with purple $Z$ being the preference, and red $\theta$ being the language model.}
%     \label{fig:markov}
% \end{figure*}

\section{Evaluation} \label{sec:exp}

In this section, we evaluate the impact of rationales on direct preference learning. We conduct multiple experiments with two main goals in mind: \textbf{(1)} to understand how the added rationales affect the efficacy and efficiency of current preference learning algorithms, and \textbf{(2)} to determine the significance of rationale quality for effective learning.

\subsection{Experimental Setup}

\textbf{Datasets.} For our analysis, we focus on three preference datasets: Orca DPO Pairs~\citep{orcadpo}, which is a pairwise preference dataset version of Orca~\citep{mukherjee2023orca}, a binarized UltraFeedback~\citep{tunstall2023zephyr}, which is a pair-wise version of UltraFeedack~\citep{cui2023ultrafeedback}, and Anthropic Helpful and Harmless, which is a huamn preference dataset about helpfulness and harmlessness. For each dataset, we take $512$ fixed samples as test set for winrate evaluations. We generate rationales and add rationales to the current datasets. We refer readers to Appendix~\ref{app:rationale generation} for details on generating rationales. Given the test sets, we sample the responses from models trained with preference learning methods and compare the performance between the models by measuring the winrates between the corresponding responses. \newline
\textbf{Models.} We investigate preference training on various large language models: Mistral-7B-v0.1, Mistral-7B-Instruct-v0.2~\citep{jiang2023mistral}, Zephyr-7B-Beta\citep{tunstall2023zephyr}, and Llama3-8B-Instruct~\citep{llama3modelcard}. We use GPT-4o~\citep{achiam2023gpt} as a judge to evaluate the responses generated by the models and to retrieve the winrate scores. Please see Appendix~\ref{app:evaluate winrate} for details. We provide full results with ablation on hyperparameters in Appendix~\ref{app:models and hyperparameters}.\newline
\textbf{Methods.} In our experiments, we study the integration of rationales into preference learning frameworks, such as DPO~\citep{rafailov2024direct}, which requires the SFT model for the reference model, and ORPO~\citep{hong2024reference} and SimPO~\citep{meng2024simpo}, which do not. To ensure fair comparison between DPO and RDPO, we fine-tune the base model with supervised fine-tuning (SFT) only using the chosen
responses from the preference dataset. We extend the code implementation from human-aware loss functions (HALOs) repository~\citep{ethayarajh2023halos} to adapt to our methodology and borrow the hyperparameters from each of the above methods in our study.

\subsubsection{Learning Dynamic with Rationales}

\begin{wrapfigure}{r}{0.65\textwidth}
% \begin{figure}[h]
% \vspace{-1.5em}
    \centering%
	\includegraphics[width=0.99\linewidth]{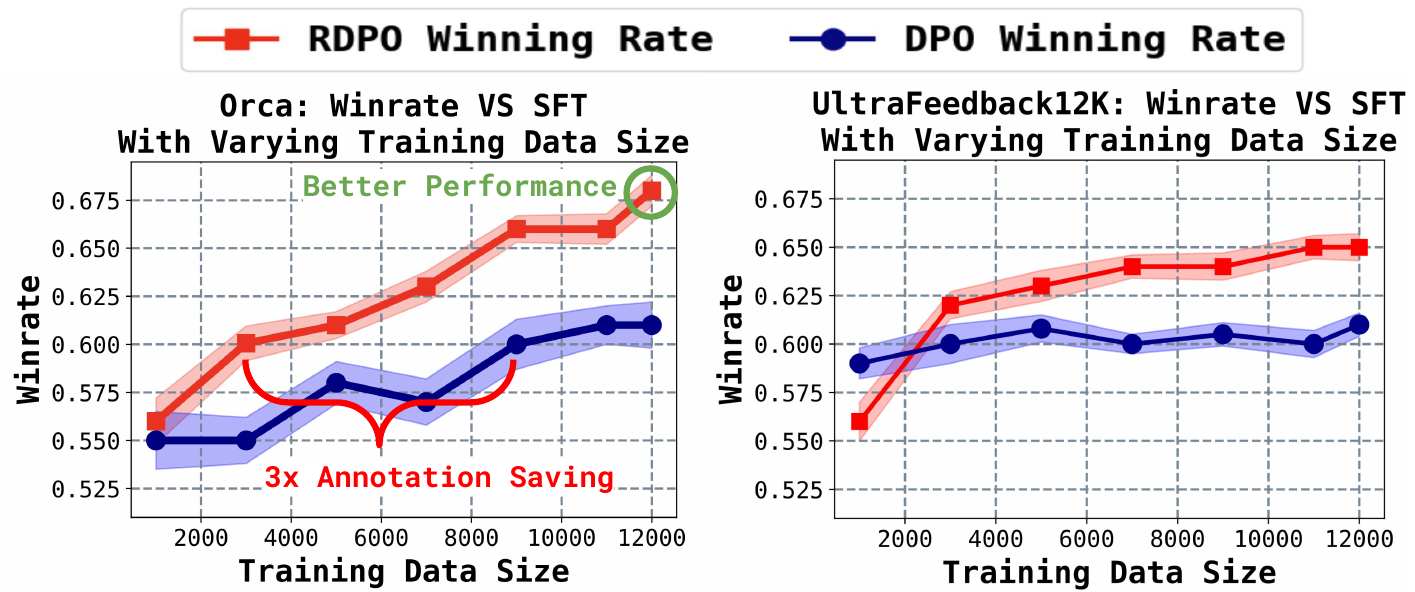}
	\caption{\update{Winrate} comparison between the models trained with RDPO and DPO. \textbf{Left:} Winrate against the SFT model trained on the Orca dataset. \textbf{Right:} Winrate against the SFT model trained on the Ultrafeedback dataset. X-axis denotes the training data size used for preference training of DPO and RDPO models. 
    % \ruoxi{remember to change "data saving" to "annotation saving" and anywhere that mentions "data efficiency" to "annotation efficiency" or "human feedback efficiency"}
    }
    \label{fig:winrate_vs_sft}
    \vspace{-1em}
% \end{figure}
\end{wrapfigure}

\paragraph{Versus SFT.} We carefully examine the dynamics of preference learning by studying how adding rationales to the current preference learning algorithms can impact performance. We compare the responses generated by the preference-aligned model against the responses generated by the SFT model and measure the winrates scores using the GPT-4o evaluator. To study data annotation efficiency, we train the models on various training data sizes, ranging from $1{,}000$ to $12{,}000$ data points, for both RDPO and DPO training. We observe on the left side of Figure~\ref{fig:winrate_vs_sft} that both models, DPO and RDPO models, achieve a better winning rate over the SFT model (more than $50\%$) on the Orca dataset with an increasing winning trend when training data increases. Additionally, we observe as the DPO model converges to around $60\%$ winning rate at the $9{,}000$ mark against the SFT model, the RDPO model achieves this rate at a smaller training data size with $3{,}000$ training data points. Furthermore, we observe that the RDPO model can reach an even higher winning rate against the SFT than the DPO model, reaching above the $66\%$ winning rate. We see a similar observation with the models trained on the UltraFeedback dataset on the right side of Figure~\ref{fig:winrate_vs_sft}. 
% Furthermore, the drawing rate for the RDPO model is stable and low across different training data sizes, which shows that RDPO winning rate is higher not due to flipping the draw points but the losing points.
While RDPO can increase the computation time due to the addition of rationales, the model trained on rationales can converge earlier with fewer data points than DPO. This is especially important as the cost of collecting human preference data is high. Thus, improving annotation efficiency can potentially save further training costs.
% \ruoxi{discuss implications. the cost of collecting human prefernce is high. requiring fewer data points means saving cost} 
Additionally, with enough computation, RDPO can reach a better model than DPO does.

Moreover, we observe for the case of the UltraFeedback dataset, with more training data, the performance of the DPO model decreases. This can be attributed to the problem of DPO overfitting and exploiting length in longer responses~\cite{azar2024general, park2024disentangling}. Indeed, the UltraFeedback dataset contains chosen responses that are longer on average ($1{,}305$ character length) than the rejected responses ($1{,}124$ character length).
% \ruoxi{can we provide some numerical evidence }, 
unlike the Orca dataset ($784$ and $978$, respectively).

% \paragraph{With enough data budget, RDPO converges to better solutions than DPO}

\begin{wrapfigure}{r}{0.6\textwidth}
% \begin{figure}[h]
% \vspace{-2.5em}
    \centering%
	\includegraphics[width=0.95\linewidth]{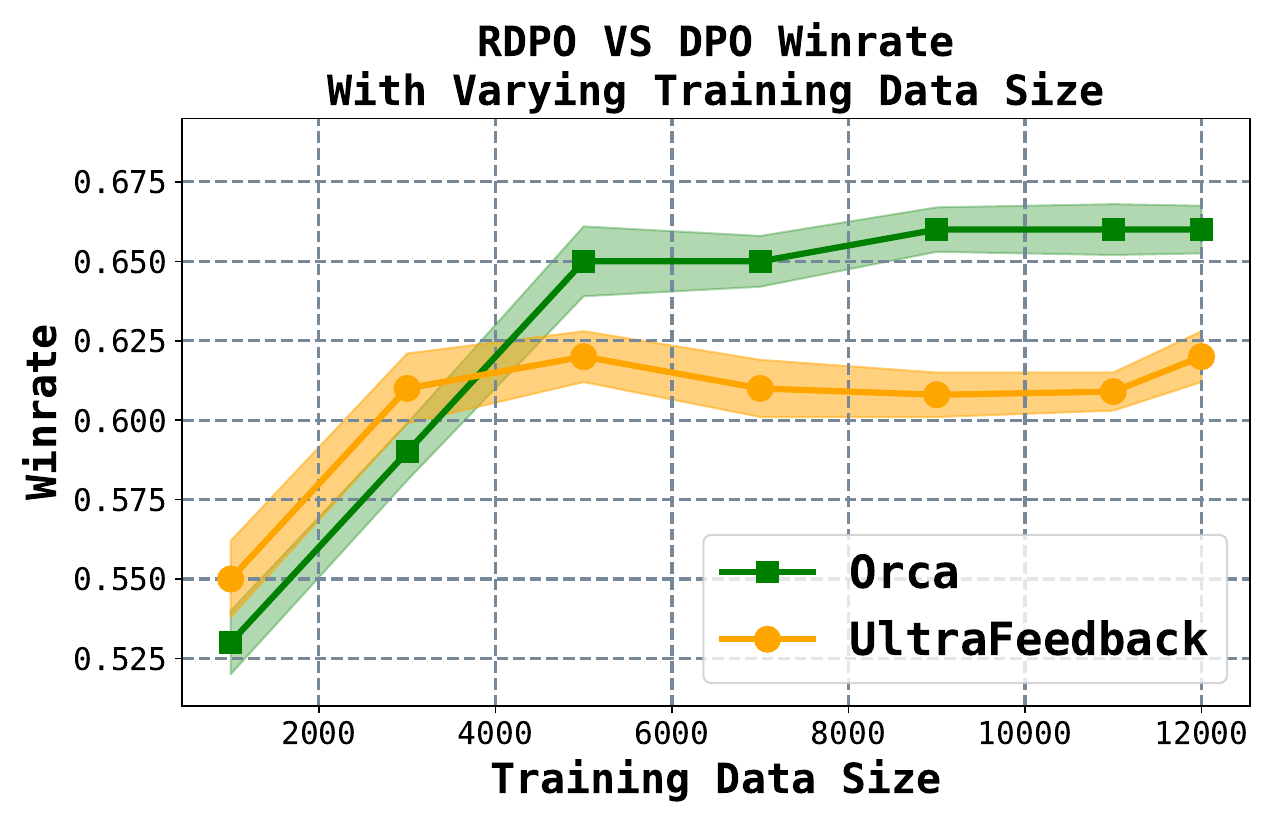}
	\caption{Winrate of RDPO model against the DPO model on respective datasets, Orca on the \textbf{left} and UltraFeedback on the \textbf{right}. The purple dashed line denotes the $0.5$ mark. }
 % \ruoxi{change data increase to data saving in the plot}
    \label{fig:winrate_vs_dpo}
% \end{figure}
\end{wrapfigure}

\paragraph{Versus DPO.} While we used SFT as a proxy to compare the performance between the DPO- and RDPO-trained models, here we directly compare the responses between these two models to measure the winrate for Orca and UltraFeedback datasets. We choose a DPO-trained model checkpoint for each dataset, where the winrate of the DPO model against the SFT model has converged, which is at $11{,}000$ and $12{,}000$, respectively. In Figure~\ref{fig:winrate_vs_dpo}, we observe that the model trained with RDPO generates better responses on average than the model with DPO, even when trained with as little as $1{,}000$ data points. With increasing training data, RDPO model improves the winrate to reach above $60\%$ in both datasets. RDPO-trained model generates more of the preferred responses than the DPO-trained model does, even with $10\times$ fewer training points.

\begin{wraptable}{r}{5.0cm}
% \begin{table}
\centering
% \vspace{-1em}
\resizebox{0.99\linewidth}{!}{
\begin{tabular}{c|ccc|c}
      ORPO         &44 & : & 56     & RORPO   
     % \toprule
     % \cmidrule[0.9pt](l{0.525em}r{0.525em}){1-1}
     % \cmidrule[0.9pt](l{0.525em}r{0.525em}){2-2}  
     %    43 & 55  \\
      % \bottomrule
\end{tabular}
}
\vspace{1em}
    \caption{Adapting rationales to the ORPO preference learning algorithm on Mistral-7B-v0.2-Instruct (Orca). Comparing the winrate of the ORPO- \textbf{(Left)} against the RORPO-trained model \textbf{(Right)}.}
    % \ruoxi{explain the shorthand RORPO somewhere}}
\label{table: orpo rorpo}
% \vspace{-1em}
% \end{table}
\end{wraptable}

\paragraph{Adaptation to ORPO.} 
% \ruoxi{To demonstrate the flexibility of our rationale-augmented learning framework}
To demonstrate the flexibility of our rationale-enriched preference learning framework, we extend the ORPO preference learning algorithm~\cite{hong2024reference}, which omits the SFT step, to include the rationales similar to the RDPO loss, and we call it RORPO. As shown in Table~\ref{table: orpo rorpo}, rationales can enhance the performance of ORPO and achieve a better winrate over the vanilla ORPO trained model. By successfully adapting rationales to both ORPO and DPO, we emphasize the simplicity of the framework as well as the effectiveness of rationales in preference learning. We further study the adaptation of these methods with rationales and evaluate the preference-trained models on the instruction-following benchmark, AlpacaEval 2.0~\citep{alpaca_eval, dubois2024length}.
% , in Appendix~\ref{app:more methods}, and we observe consistent results when evaluating on this benchmark.

% \subsection{Evaluation of the Impact of Rationales with AlpacaEval 2.0}\label{app:more methods}

\paragraph{Evaluation of the Impact of Rationales with AlpacaEval 2.0.} Here, we study the flexibility of our method by extending different preference learning methods, such as DPO~\citep{rafailov2024direct} and ORPO~\citep{hong2024reference}, with rationales. In this experiment, we evaluate the performance of the trained models on the automatic instruction-following AlpacaEval 2.0 benchmark~\citep{alpaca_eval} with GPT-4-turbo as a judge and report the raw winrate, the length-controlled (LC) winrate~\citep{dubois2024length}, which is robust against the verbosity bias that the raw winrate inherently entails, and the average response length. We train the instruction-tuned models, Mistral-7B-Instruct-v0.2 and Llama-3-8B-Instruct, on the Intel-DPO-Pairs dataset. 

% \begin{figure}[h]
%   \begin{center}
%     \includegraphics[width=0.9\linewidth]{fig/alpaca_dpo_rdpo.png}
%   \end{center}
%   \caption{The performance comparison of the original model, DPO trained model, and DPO with rationales (RDPO) trained model on the AlpacaEval 2.0 benchmark. The \textbf{bolded} numbers denote the average response length of each models.}
%   % \ruoxi{shall we move it to intro}}
%     \label{fig:alpaca dpo and rdpo}
% \end{figure}

% \begin{figure}[h]
%   \begin{center}
%     \includegraphics[width=0.9\linewidth]{fig/alpaca_orpo_rorpo.png}
%   \end{center}
%   \caption{The performance comparison of the original model, ORPO trained model, and ORPO with rationales (RORPO) trained model on the AlpacaEval 2.0 benchmark. The \textbf{bolded} numbers denote the average response length of each models.
% }
%   % \ruoxi{shall we move it to intro}}
%     \label{fig:alpaca orpo and rorpo}
% \end{figure}

\begin{wraptable}{r}{8.0cm}
% \begin{table}[]
\resizebox{1.0\linewidth}{!}{
\begin{tabular}{l|lll|lll}
         & \multicolumn{3}{l|}{Mistral-7B-Instruct-v0.2} & \multicolumn{3}{l}{Llama-3.1-8B-Instruct} \\
         
         & LC WR      & Winrate      & Length      & LC WR     & Winrate     & Length     \\
         \toprule
Original & 17.11           & 14.72        & 1676        & 22.92          & 22.57       & 1899       \\
\midrule
DPO      & 19.52           & 15.81        & 1632        & 26.02          & 22.52       & 1759       \\
RDPO     & 22.42           & 17.87        & 1627        & 27.55          & 23.69       & 1750       \\
\midrule
ORPO     & 12.84           & 12.24        & 1782        & 23.11          & 20.27       & 1734       \\
RORPO    & 20.45           & 16.27        & 1618        & 26.55          & 23.45       & 1758      
\end{tabular}
}
\caption{The performance comparison of the original model, ORPO trained model, and DPO/ORPO with rationales (RDPO/RORPO) trained model on the AlpacaEval 2.0 benchmark.}
\label{fig:alpaca}
% \end{table}
\end{wraptable}

We observe in Figure~\ref{fig:alpaca} that DPO trained model improves the winrates on both models compared to the original model. Additionally, RDPO models further increase the win rates. We note that in the case of the Mistral model, the winrate decrease with ORPO preference training, which might be due to the lack of the reliance on the reference model, the behavior which is also observed in~\citet{ethayarajhmodel}. Furthermore, after adding rationales, the (LC) winrate not only increases but also surpasses of the original model. These results show the helpfulness of adding rationales into preference learning.
Interestingly, we also note that rationale based models increase the winrates while their average response lengths are decreased compared to average response lengths of the original model, which is not a similar observation as seen in some current methods, such as SPPO~\citep{wu2024self} or SimPO~\citep{meng2024simpo}.

\begin{wraptable}{r}{6.0cm}
% \begin{table}[]
\resizebox{1.0\linewidth}{!}{
\begin{tabular}{lcc|cc}
      LLaMA-3.1-8B-I & \multicolumn{2}{l}{UltraFeedack} & \multicolumn{2}{l}{Anthropic HH} \\
       \toprule
   Dataset Size    & 20K             & 40K            & 20K             & 40K            \\
RSimPO Winrate & 61              & 68             & 58              & 65            
\end{tabular}
}
\caption{The winrate of the RSimPO trained model on 20K and 40K dataset versus the SimPO trained model on the full 40K dataset.}
\label{table: rsimpo}
% \end{table}
\end{wraptable}

\paragraph{SimPO: Full Size Dataset} Here, we study the effectiveness of training with rationales on the full size dataset, and we adapt SimPO to incorporate rationales, and train the RSimPO model, accordingly. In particular, we train the LLaMA-3.1-8B-Instruct model on the UltraFeedback dataset and Anthropic HH dataset. After proper filtering, we are left with around 40K samples for each dataset. To show the effectiveness of RSimPO, we observe in Table~\ref{table: rsimpo} that RSimPO achieves a higher winrate against the SimPO trained model, achieving over 65\% winrate on both datasets agains the SimPO model. Furthermore, to show the efficacy of RSimPO, we train the model on half of the dataset, and we still observe better performance compared to the fully trained SimPO model on 40K for both data sets.

% \paragraph{ SimPO: Full Size Dataset} Here, we study full size dataset

\subsection{Investigation of the Impact of Rationales}
\label{sec:rationale_impact}

% \update{\subsubsection{Rationale-Only Optimization}}
% \label{app:ref only}

\textbf{Rationale-Only Optimization.} We study an the impact of the introduction of rationale term into the preference learning. To address this, we conducted a series of experiments to isolate the impact of each component and evaluate their combined effect. Specifically, we investigated an extreme case where the rationale loss alone drives preference optimization, with the DPO alignment loss set to zero. This approach was based on the hypothesis that rationales inherently encode preferences by combining preference-response pairs, the preferences themselves, and the associated reasoning processes, thereby providing a rich and effective training signal.
For these experiments, we fine-tuned Mistral-7B-Instruct-v0.2 on the Orca dataset across three settings: RDPO (combining DPO and rationale loss), DPO (excluding rationale loss), and Rationale-Only (excluding DPO loss). The results, as shown in the Table~\ref{tab:rationale only}, reveal that rationales alone can substantially improve model performance, achieving a high win rate of over 61\% without explicit pairwise preference modeling. This improvement likely stems from the informational richness embedded in rationales, which compensates for the absence of pairwise alignment.

\begin{wraptable}{r}{6.0cm}
% \begin{table}[h!]
% \vspace{-1em}
\centering
\resizebox{0.90\linewidth}{!}{
\begin{tabular}{cccc}
      
         & RDPO & DPO & Rationale-Only      \\
         \cmidrule[0.9pt](l{0.525em}r{0.525em}){2-4}
General  & 64.5       & 59.1  &    61.8  \\
Detailed & 64.4      & 59.1  & 61.3   \\
\midrule

\end{tabular}
}
\caption{The impact of different components of RDPO by measuring the winrate of the target model against the SFT model. The results on Mistral-7B-v0.2-Instruct model using the Orca dataset. }
\label{tab:rationale only}
% \end{table}
\end{wraptable}
While DPO also demonstrated a majority win rate against the SFT baseline, training with both rationale and preference losses (RDPO) consistently achieves the highest win rate (64.5\%) across both general and detailed settings. This highlights the benefit of integrating rationales into the preference objective, effectively leveraging the strengths of both losses to produce superior performance.
% To further investigate how rationales enhance DPO preference learning, we examined the reward margin metrics. As shown in the Table~\ref{tab:reward margin}, RDPO not only achieved higher reward margins between chosen and rejected responses but also demonstrated faster convergence compared to DPO. This can be explained through the following: while DPO explicitly aims to maximize reward margins, the inclusion of rationales provides an implicit quality signal, offering explanations for the differences between chosen and rejected responses. This signal reinforces the model's ability to improve reward margins by guiding it toward more informed preferences.

% \begin{table}[h!]
% \centering
% \resizebox{1.0\linewidth}{!}{

% \begin{tabular}{ccccccccccccc}
% % \toprule

% \textbf{Training Points} & 0 & 1000 & 2000 & 3000 & 4000 & 5000 & 6000 & 7000 & 8000 & 9000 & 10000 & 11000 \\
% \toprule
% \textbf{DPO} & 0.00 & 0.05 & 0.19 & 0.32 & 0.42 & 0.49 & 0.54 & 0.58 & 0.62 & 0.63 & 0.65 & 0.66 \\
% \midrule
% \textbf{RDPO} & 0.00 & 0.10 & 0.25 & 0.46 & 0.67 & 0.76 & 0.83 & 0.85 & 0.86 & 0.87 & 0.89 & 0.91 \\
% % \hline
% \end{tabular}
% }
% \caption{The eval reward margin comparison for different training losses, DPO and RDPO. The Mistral-7B-Instruct-v0.2 is trained on the Orca dataset. }
% \label{tab:reward margin}
% \end{table}

These findings underscore the complementary nature of the rationale SFT loss and the pairwise alignment loss. While DPO explicitly optimizes reward margins, the rationale prediction loss provides supplementary supervision, enabling the model to learn the reasoning underlying response preferences. This integration not only strengthens the selection process but also accelerates training convergence. By combining these two approaches, RDPO amplifies their individual strengths, resulting in more efficient and effective preference learning.

\textbf{Log Probabilities.} To further investigate how rationales may boost preference optimization, we take a deeper look into log probabilities of the preference pairs. Specifically, we take a look at the difference of log probabilities between the chosen and rejected response on RDPO trained Llama-3.1-8B-Instruct model with Orca dataset.

% \begin{wrapfigure}{r}{0.40\textwidth}
\begin{figure}[h]
  \begin{center}
    \includegraphics[width=0.59\linewidth]{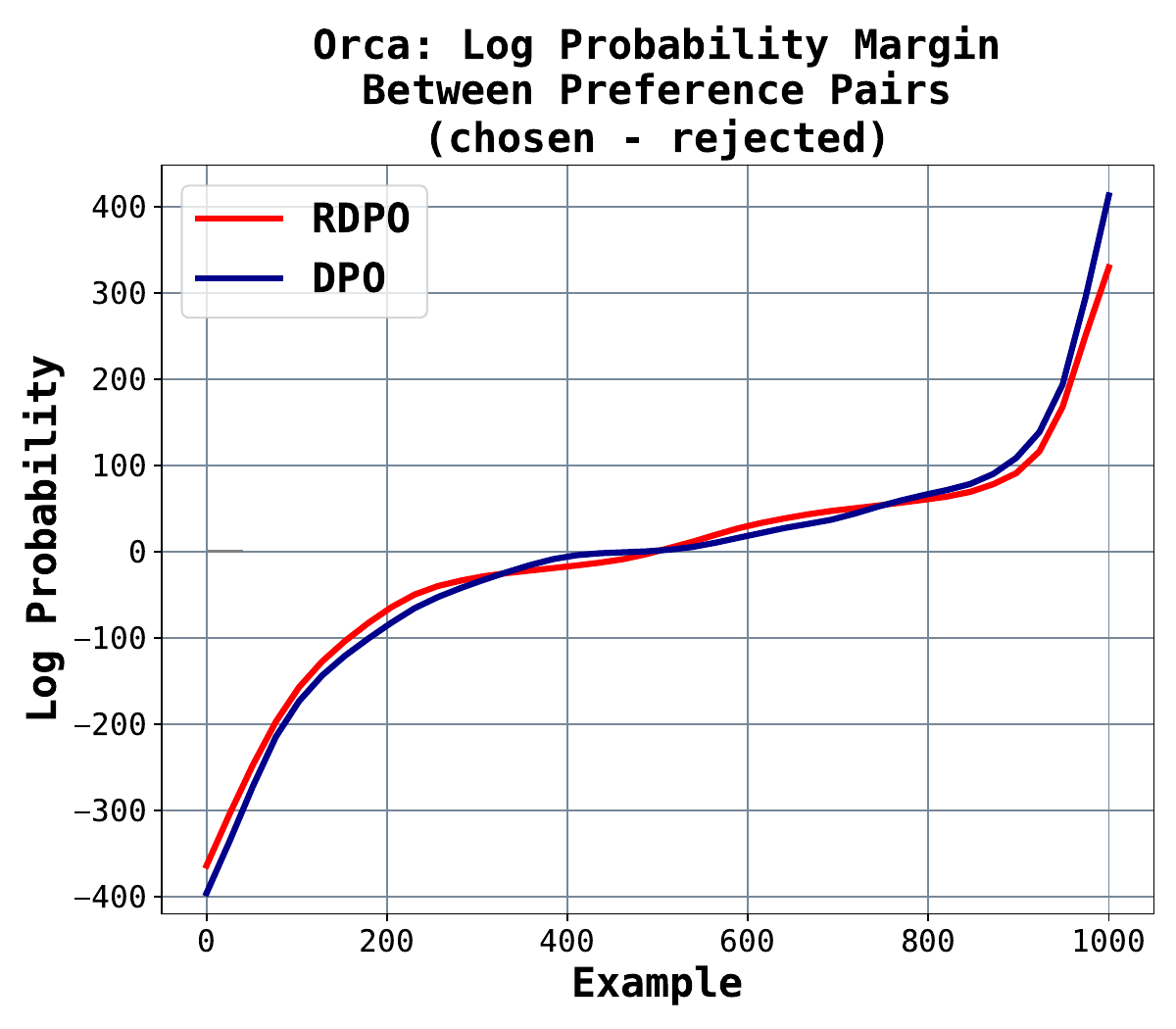}
  \end{center}
  \caption{The performance comparison of the original model, ORPO trained model, and ORPO with rationales (RORPO) trained model on the AlpacaEval 2.0 benchmark. The \textbf{bolded} numbers denote the average response length of each models.
}
  % \ruoxi{shall we move it to intro}}
    \label{fig:log margin}
\end{figure}
% \end{wrapfigure}
As shown in Figure~\ref{fig:log margin}, at the tails of the plots, where the margins are large, DPO yields higher margins than RDPO. This suggests that DPO may be exploiting certain biases or engaging in reward hacking to maximize reward margins. In contrast, RDPO appears to act as a regularizer, mitigating such exploitative behavior. Interestingly, at the 0 level—where the margins are negligible, indicating that preference pair responses are difficult to differentiate, DPO remains relatively flat, whereas RDPO crosses the 0 line at a steeper angle. This suggests that RDPO is more effective at handling and resolving ambiguous cases compared to DPO. The investigation into the impact of rationales on preference learning suggests that they may serve as valuable cues, potentially acting as a regularizer to enhance preference learning.

\textbf{Quality of Rationales.} To investigate the quality of generated rationales, we incorporated the more capable LLM, GPT-4o, to evaluate the generated rationales in terms of helpfulness, relatedness, correctness, and coherence. We further examine the importance of the quality of rationales for preference learning in Appendix~\ref{sec:quality_analysis}, studying how different types of rationales would impact preference learning.

\section{A Simplified Theoretical Model for Rationales}
% \hoang{I moved the earlier analysis to Appendix~\ref{app: theory} for now.}
To better understand rationale-enhanced preference learning, we employ machinaries from information theory to quantify benefits rationales provide in learning ground truth preferences under simplified assumptions. Formally, given query $X$, let the preference $Z$ be a binary random variable, with $Z = 1$ indicating that response $Y_1$ is preferred over response $Y_2$, and $Z = 0$ indicating the opposite. Let $R$ denote the rationale, $S = (X, Y_1, Y_2)$, and assume that the dataset $D=\{S_i,R_i,Z_i\}_{i=1}^n\sim \mu^n$  is sampled i.i.d. from a distribution $\mu$. 
% We denote the dataset with rationales removed as $D_{\setminus R}$. 
% Throughout this section, we use standard quantities, such as the entropy $H(\cdot)$, the mutual information $I(\cdot ; \cdot)$, and their conditional counterparts $H(\cdot | \cdot)$ and $I(\cdot ; \cdot | \cdot)$. We define $\mathcal{A}_{(\cdot)}(D)$ as the learning algorithm applied to the dataset $D$.

As an intuitive first step, we quantify the benefits using the conditional mutual information between the true preferences and rationale-implied preferences given the input-response pairs, i.e., $I(Z;g(R)|S)$, where $g(R)$ capture the preference inferred from the rationale $R$. 
Intuitively, this mutual information quantity characterizes the added value of rationales in understanding true preferences, measuring how much additional information rationales provide beyond what can be inferred from input-response pairs $S$. Our analysis demonstrates a closed-form relationship between rationale informativeness and its alignment with true preferences (see Appendix~\ref{app:MIanalysis} for detailed derivations of all results). 
% This theoretical insight aligns conceptually with our experimental observations in Section~\ref{sec:quality_analysis}: when we manually inject noise into rationales to misalign them with true preferences, we observe hampered preference learning, as indicated by lower winrates. It's worth noting that while this connection is clear, the experimental setting involves additional complexities not captured in our simplified model (see Appendix~\ref{app:MIanalysis} for detailed derivations of all results).

Next, we quantify the benefits provided by rationales by directly analyzing the generalization error for preference learning with and without rationales. Compared with analyzing the informativeness of rationales outlined above, this approach offers a more direct measure of the impact of rationales on learning outcomes. Note that the generalization error of preference learning still differs from the winrate metrics used in our experiemnts for evaluating the generation of preference-aligned models. However, they are intuitively connected under the DPO loss: the minimized test loss also indicates learning the optimal data generation policy that achieves the highest reward. We defer the detailed statement of the theorem and proof to Appendix~\ref{app:pfthm1}. In particular, our theoretical derivation shows that training with rationales can lead to improved lower generalization error when the rationale does not contain irrelevant information other than those predictive of the preference $Z$, and the learning process only captures the rationale information that is useful to predict $Z$. Despite being derived from a simplified model, our theoretical insights align well with experimental observations. For instance, our evaluation in Section~\ref{sec:quality_analysis} demonstrates that detailed rationales, which may contain extraneous information, achieve lower sample efficiency compared to more general rationales that focus on preference-predictive content; furthermore, when we manually inject noise into rationales to misalign them with true preferences, we observe hampered preference learning, as indicated by lower winrates.
% We consider two regimes: 
% \emph{1)} \textbf{Training with rationale}: Let $\theta_{\text{ra}}=\mathcal{A}_{\text{ra}}(D)\sim P_{\theta_{\text{ra}}|D}$ denote the parameters of the language model trained to predict $Z$ given $S$ and $R$. \emph{2)} \textbf{Training without rationale}: Let $\theta_{\text{un}}=\mathcal{A}_{\text{un}}(D_{\setminus R})\sim P_{\theta_{\text{un}}|D_{\setminus R}}$ denote the output parameters trained to predict $Z$ given only $S$. 

% First, we analyze the conditional mutual information $I(Z; R | S)$, which measures the additional information that the rationale $R$ provides about the preference $Z$ beyond what the query and responses alone provide. Our result identifies three distinct regimes of rationale informativeness: uninformative, maximally informative, and moderately informative. The informativeness is most pronounced when the rationale is highly predictive of the preference and prediction based on the query and responses alone tends to be biased. The detailed expression for the mutual information and the proof can be found in Appendix~\ref{app:MIanalysis}.

\section{Conclusion \& Limitations} \label{sec: conclusion}

In this work, we explored and validated a data-centric perspective on direct preference learning for aligning language models. Standard approaches train models on preference pairs (preferred vs. dispreferred responses), but often lack explicit information about why one response is superior, leading to ambiguity that can hinder learning efficiency and require substantial, costly datasets. To address this limitation and extract more value from existing human preference data, we introduced the concept of augmenting these pairs with rationales, which explain the reasoning behind each preference. We presented a straightforward framework for integrating readily available, machine-generated rationales into existing direct preference optimization pipelines. Our extensive empirical evaluations demonstrate that this rationale-augmented approach significantly enhances the learning process: it increase the effective use of each annotated data point, improves model convergence to higher performance levels, and helps mitigate undesirable behaviors. Furthermore, theoretical insights grounded in information theory corroborate the observed efficiency gains provided by rationales. Ultimately, our findings underscore the significant potential of thoughtfully designing the structure of preference data itself. Enhancing existing datasets with explanatory rationales offers a powerful and complementary strategy alongside algorithmic advancements for achieving more robust and efficient language model alignment.
% In our work, we propose a paradigm shift in preference learning, emphasizing a data-centric perspective. Language models are trained by presenting them with pairs of answers, one preferred and one dispreferred, with the objective of teaching them human preferences. However, the selection of preferred answers can often be ambiguous without explanation. To address this challenge and enhance preference optimization efficiency, we introduce rationales that provide explicit reasoning for choosing one answer over another. We propose a straightforward adaptation of existing losses by incorporating these rationales into the training pipeline. Through extensive empirical experiments, we demonstrate that rationales significantly enhance training data annotation efficiency and lead to improved performance compared to baseline methods. Moreover, our approach is grounded in information theory, offering insights into how rationales enhance preference training efficiency.
To date, we have integrated rationales into our training process and successfully trained models with up to 8 billion parameters. We encourage further investigation into the impact of rationales on preference learning, particularly exploring larger models and datasets. To facilitate research in this area, we make our code and datasets publicly available. With the development of unpaired preference learning algorithms, such as KTO~\cite{ethayarajh2024kto}, it is important to extend the use of rationales to handle unpaired responses in future work, e.g., such as provided in the UltraFeedback dataset~\cite{cui2023ultrafeedback} contains rationales for single responses without pairwise comparison. We also would like to study the impact of rationales on trained the reward models, which might have a different impact than the critiques from existing literature.

% Getting high-quality data evaluated by hu

\newpage

\section*{Impact Statement}

This paper presents a data-centric extension aimed at improving preference learning by enabling users to provide rationales for their choices, particularly in cases where such reasoning is not easily inferred by the model. By granting users greater control over data creation, this approach helps guide the model in effectively capturing user-specific preferences. Through responsible data curation and generation, this method has the potential to significantly influence model behavior in real-time applications.

\section*{Acknowledgments}

Ruoxi Jia and the ReDS lab acknowledge support through grants from the Amazon-Virginia Tech
Initiative for Efficient and Robust Machine Learning, the Cisco Award, the Commonwealth Cyber Initiative Cybersecurity Research Award, the National Science Foundation under grants CNS-2424127, IIS-2312794, IIS-2313130, OAC-2239622, and OpenAI API research credits.

\bibliography{colm2025_conference}
\bibliographystyle{colm2025_conference}

\appendix
\newpage
\clearpage

\appendix
\begin{appendices}{}

% \addcontentsline{toc}{section}{Appendix} % Add the appendix text to the document TOC
% \part{Appendix} % Start the appendix part
% \parttoc % Insert the appendix TOC

\startcontents[sections]
\printcontents[sections]{l}{1}{\setcounter{tocdepth}{2}}

\onecolumn

\section{Extended Related Work} \label{app: ext rel work}

% Method
% Problems
% Solutions

\paragraph{RLHF with Reward Modeling.} Tuning large language models to align their outputs towards human preferences is crucial for controlling model behavior and maintaining desirable boundaries~\citep{casper2023open}. To achieve this, RLHF has been introduced; aligning models through preference training~\citep{christiano2017deep,ziegler2019fine,stiennon2020learning}. \citet{schulman2017proximal} describe a method that typically involves two stages. The first stage learns a reward model using a preference dataset often modeled under the Bradley-Terry model~\cite{bradley1952rank}. The second stage fine-tunes the target policy model to maximize the rewards from the reward model, employing algorithms such as proximal policy optimization (PPO) proposed by \citet{schulman2017proximal} and adopted in \citet{ouyang2022training}. A direct preference optimization (DPO) method that implicitly models the reward function was introduced by \citet{rafailov2024direct}. However, \citet{azar2024general} observe that RLHF and DPO are prone to overfitting due to the assumptions of the Bradley-Terry model. Conversely, \citet{pal2024smaug} explore the possibility of DPO underfitting when dealing with challenging responses that are difficult for the model to distinguish. Additionally, \citet{park2024disentangling} note that DPO can exploit response length to maximize reward, proposing a length-regularized DPO (R-DPO) to address this issue. This should not be confused with our rationale-based DPO (RDPO) method. Interestingly, we observe that if rationales mention conciseness as a feature, then the length of responses is significantly reduced compared to SFT and DPO responses. The learning dynamics during preference tuning are analyzed in \citet{im2024understanding}, emphasizing the importance of high-quality preference datasets for effective learning. They find that the more distinguishable the response pairs, the easier it is for the model to learn, leading to faster convergence. This has also been observed in \citet{pal2024smaug}. However, designing such datasets is challenging, and it also remains important for models to learn from such nuanced responses, which appear in practice. We try to address the difficulty of model learning from intricate preferences by providing rationales during preference training. To improve efficiency over DPO, which requires an intermediate step to train the reference model, odds ratio
preference optimization (ORPO) was introduced by \citet{hong2024reference} to eliminate this step. Another method by \citet{ethayarajh2024kto} adapts the Kahneman-Tversky human utility model to handle preference datasets with a single response (either chosen or rejected), removing the need for training the model on both responses. Conversely, \citet{yuan2023rrhf} propose a preference method that considers multiple ranked responses for a prompt and optimizes over them. Our method can complement these methods by adding rationales into training. In this paper, we demonstrate an extension of our framework to ORPO.
% \ruoxi{comment on how our method differs from or complements existing work throughout this section and point out any interesting connections (like sharing similar or opposite findings, etc.)}
% KTO 
% Ranking
% Another type of methods
% schulman2017proximal
% \citep{rafailov2024direct}

\textbf{General Preference Modeling.} 
Reward modeling, however, can incentivize undesirable behaviors, such as ``reward hacking''~\citep{amodei2016concrete}, where agents maximize rewards without achieving the desired objective. Overfitting is another challenge, as exemplified in~\cite{azar2024general}. While effective for comparing two responses, the Bradley-Terry preference modeling relies on the assumption of transitivity, which may not hold true in practice~\citep{bertrand2023limitations}. To address this,~\citet{munos2023nash} introduced general preference modeling, which directly learns general preferences by formulating a two-player, constant-sum game between policies. The goal is to maximize the probability of generating the preferred response against the opponent. The solution is the Nash equilibrium of this game, where payoffs are derived from the general preference function. Building upon this work,~\citet{munos2023nash} proposed an algorithm for the regularized general preference model, while~\citet{swamy2024minimaximalist} developed a solution for the unregularized formulation and introduced self-play preference optimization (SPO) as an iterative algorithm to reach the optimal solution. However, SPO suffers from data inefficiency due to its two-timescale update rules. To address this,~\citet{rosset2024direct} introduced an efficient direct Nash optimization (DNO) method that leverages the DPO formulation in practice. Additionally,~\citet{wu2024self} proposed an efficient, scalable, iterative self-play method that generates responses generally preferred over others.

While previous efforts have introduced algorithmic enhancements for preference tuning, they have been limited to the existing framework of preference datasets with prompts and ranked responses. In contrast, our work is first to introduce rationales, a data-centric solution, into preference learning.
\newline
\textbf{Learning with Rationales.} 
The supervised learning framework typically involves training a model to learn the ground truth label for a given prompt without providing explicit explanations for the associations, which can lead to the model learning incorrect cues. To mitigate this issue, rationales have been integrated into the framework, offering explanations for the given associations. These rationales initially were generated by humans~\citep{zaidan2007using, ross2017right, hase2021can, pruthi2022evaluating}. However, due to the high cost of human labor and the development of more capable large language models, rationales are now often automatically generated by these models, reducing the need for human involvement~\citep{wei2022chain, kojima2022large}. Given rationales, they have been used as guiding aids by incorporating them directly into the prompt during the training phase~\citep{rajani2019explain, zelikman2022star, huang2023large} or at the inference stage~\citep{wei2022chain, kojima2022large, wang2022self}. Besides using them as additional context within the prompt, rationales can also serve as labels to train models to generate such explanations for their predictions~\citep{wiegreffe2021measuring, narang2020wt5, eisenstein2022honest, wangpinto, ho2023large, magister2022teaching, li2023symbolic}. In similar manner, rationales have been applied in knowledge distillation, where they are generated by a more capable models to supervise weaker models~\citep{hsieh2023distilling, chen2024learning}. In parallel with these advancements, we introduce rationales into the preference learning landscape, where rationales are used to explain the preference of one answer over another. Our findings demonstrate the effectiveness of rationales in preference learning, even when generated by the same model or a smaller-sized model.

\update{\textbf{Synthetic Preference Data Generation.}} Synthetic preference data generation plays a pivotal role in preference learning by creating new annotated datasets that capture user choices or preferences~\cite{yangrlcd, pace2024west, meng2024simpo}. These methods focus on producing preference pairs that can serve as training data for models, enabling the exploration of diverse scenarios and reducing reliance on costly manual annotations. Furthermore,~\cite{wu2024self, wang2024self, meng2024simpo} try to further synthesize the preference examples by iteratively generating new response pairs in on-policy manner. However, our approach diverges fundamentally from this objective. While synthetic data generation targets the creation of new datasets, our work emphasizes enhancing existing, fixed datasets with provided preferences pairs by incorporating rationales, thereby enriching preference annotations with explanatory depth. This distinction highlights the complementary nature of these methods: data generation addresses the early stage of creating foundational datasets, whereas rationale augmentation enhances the interpretability and utility of existing data. Attempting to compare these approaches directly would obscure their unique contributions. Instead, their synergy lies in how synthetic data generation can produce the raw preference pairs that are later refined through rationale augmentation, advancing the overall preference learning pipeline. Exploring how these two approaches can be combined to create synthetic datasets enriched with rationales is an exciting direction for future work, holding promise to further enhance the capabilities and generalizability of preference learning models.

% avoid reward hacking amodei2016concrete

% % \paragraph{Information Theory.} 

% general theoretical azar
% smaug 

% orpo

% len dpo

% ranking preference learning

% general preference learning
% nash learning from human feedback
% direct nash optimization
% self play yue wu

% understanding the learning dynamics of alignment with human feedback shawn Im

% In our study, we show the impact of incorporating rationales
% By incorporating rationales, we enhance performance and improve data efficiency, marking a substantial advancement in the field. \ruoxi{the last sentencce carries exagerative and subjective tone. remove or revise to make it more objective. as we didn't compare with many of the works mentioned here, it is important to carefully justify why we didn't compare, e.g., our efforts are complementary to exisitng work}

\newpage

\section{Theoretical Derivations} \label{app: theory}

We begin with defining several standard quantities to be used throughout this section.

\begin{definition}
    Let $X,Y$ and $Z$ be arbitrary random variables, and let $D_{KL}$ denote the KL divergence. We denote $P_X$ as the marginal probability distribution of $X$, and $P_{Y|X}$ as the conditional distribution.
    
    The entropy of $X$ is given by: $$H(X) = -\sum_x P(X=x)\log P(X=x).$$
    If $X$ is a binary variable with $p=P(X=1)=1-P(X=0)$, then we use $H(p)$ for $H(X)$.

    The joint entropy of two random variables, $H(X,Y)$, is the entropy of their joint distribution.

    The conditional entropy of $X$ given $Y$, $H(X|Y)$, is: $$H(X|Y)=H(X,Y)-H(Y).$$

    The mutual information between $X$ and $Y$ is: $$I(X;Y)=D_{KL}(P_{X,Y}\|P_XP_Y).$$

    The disintegrated mutual information between $X$ and $Y$ given $Z$ is: $$I^Z(X;Y)=D_{KL}(P_{X,Y|Z}\|P_{X|Z}P_{Y|Z}).$$

    The corresponding conditional mutual information is given by:  $$I(X;Y|Z)=\mathbb{E}_Z[I^Z(X;Y)].$$

    If all entropies involved are finite, it can be shown that $I(X;Y)=H(Y)-H(Y|X)$.
\end{definition}

\subsection{Mutual information analysis}
\label{app:MIanalysis}

% \anit{This is confusing. What sort of transformation of R are we talking about here. Think of an example to provide some intuition here. Worth reminding the reader here that, there can be cases where R=1, Z=0 and R=0,Z=1 are possible as R can be potentially noisy.}
% In the rest of the text, we will overload the notation $R$ for both representing the rationale and rationale-implied preference.

Formally, given query $X$, let the preference $Z$ is a binary random variable, with $Z = 1$ indicating that response $Y1$ is preferred over response $Y2$, and $Z = 0$ indicating the opposite.  Assume that the rationale-implied preference $R$ is a binary random variable, with $R = 1$ indicating a rationale that supports $Y1$ being preferred, and $R = 0$ otherwise. For example, if the rationale mentions that $Y1$ is more concise and informative than $Y2$, then $R = 1$. However, there can be cases where $R \neq Z$, as the rationale may not always align perfectly with the actual preference.

For the analysis, we consider the following model: 1) The rationale $R$ depends on the query-response (QR) pair $S=(X,Y1,Y2)$ and the preference $Z$, and is characterized by parameters $\beta$ and $\alpha$, where: $\beta = P(R = 1 | Z = 1, S) = P(R = 0 | Z = 0, S)$ represents the precision rate of consistency, and $\alpha = P(R = 1 | Z = 0, S) = P(R = 0 | Z = 1, S)$ represents the recall error due to inconsistency. 2) The preference $Z$ is modeled as $P(Z = 1 | S) = f(S) + \epsilon$, where $f(S)$ captures the preference based on the observable query-response pair $S$, and the additive noise term $\epsilon$ is a simple way to account for unobserved factors influencing the complex human preference that are not captured in $S$. \textit{The term $\epsilon$ is referred to as ``bias'' in the \textbf{main text} that accounts for the difference between the true preference and prediction based on the query and responses alone.}
% \anit{We need to provide some rationale/intuition as to why is Z modeled as such especially the additive noise term.}

\begin{theorem}
\label{thm:mutual_info}
    Under the given assumptions, the mutual information $I(Z; R | S)$ is given by:
\begin{align*}
H(p + \epsilon) 
                     &- (\beta (p + \epsilon) + \alpha (1 - p - \epsilon))\cdot H\left(\frac{\beta (p + \epsilon)}{\beta (p + \epsilon) + \alpha (1 - p - \epsilon)}\right) \\
                     &- (1 - (\beta (p + \epsilon) + \alpha (1 - p - \epsilon))) \cdot H\left(\frac{\alpha (p + \epsilon)}{\alpha (p + \epsilon) + \beta (1 - p - \epsilon)}\right),
\end{align*}
where $p=f(S)$. 
The mutual information $I(Z; R | S) $ satisfies the following properties in three distinct regimes:

1. Uninformative rationale regime: If $\beta = \alpha = 0.5$, then $I(Z; R | S) = 0$.

2. Maximally informative rationale regime: If $\beta = 1$ and $\alpha = 0$, then $I(Z; R | S) = H(p + \epsilon)$.

3. Moderately informative rationale regime: If $\beta = 0.5 + \gamma$ and $\alpha = 0.5 - \gamma$, where $0 < \gamma < 0.5$, then $I(Z; R | S)$ increases with $\gamma$, ranging from 0 when $\gamma = 0$ (uninformative rationale) to $H(p + \epsilon)$ when $\gamma = 0.5$ (maximally informative rationale).
\end{theorem}

% We delegate the detailed proof to Appendix \ref{app:miproof}.

The theorem highlights that the potential benefits of including rationales in the training process for preference learning tasks may vary in different regimes.

\textbf{Regime 1: Highly informative rationale ($\beta = 1$ and $\alpha = 0$)}: In this case, the mutual information is solely determined by the entropy of the preference prediction from the query and responses, $H(f(S) + \epsilon)$. Let us interpret $f(S)$ as the query-response-dependent (QR-dependent) confidence generator that only depends on $S$, and $\epsilon$ captures the idiosyncrasies such as unknown confounding factors that influence the probability of preference. If the true probability $P(Z = 1 | S) = f(S) + \epsilon$ is less than 0.5, i.e., $Z$ is most likely to be 0, a positive $\epsilon > 0$ means that the true probability $P(Z = 1 | S)$ is less extreme than the confidence score $f(S)$ as it gets closer to 0.5 with increasing $\epsilon$. On the contrary, a negative $\epsilon < 0$ means that the true probability $P(Z = 1 | S)$ is more extreme than the QR-dependent confidence $f(S)$ as it gets closer to 0 with increasing $|\epsilon|$.

From the perspective of the QR-dependent confidence generator $f(S)$ that tries to explain the preference based on QR pairs, a positive $\epsilon > 0$ would make it look more confident than it should be, i.e., overconfident, while a negative $\epsilon < 0$ would make it less confident (or more conservative) based on the QR sequence than it should be.

If it is likely to be overconfident based on the QR pairs, i.e., $\epsilon > 0$, then the more positive $\epsilon$ is, the more risk there is of being overconfident in the QR-dependent predictor $f(S)$. In this case, there is a lot of mutual information in $I(Z; R | S)$, so having rationales can ``soften'' the potential overconfidence by bringing additional information other than the QR pair, which would otherwise occur in QR pairs alone, as in traditional reward modeling. Similar analysis holds when $P(Z = 1 | S) = f(S) + \epsilon$ is greater than 0.5, i.e., $Z$ is most likely to be 1.

\emph{Key message:} Rationales are most useful when the reward modeling based on QR alone tends to have bias (i.e., overconfident).

\textbf{Regime 2: Uninformative rationale ($\beta = \alpha = 0.5$)}: In this regime, the rationale provides no \emph{additional} information about the preference,  and the mutual information $I(Z; R | S)$ is zero.

\textbf{Regime 3: Moderately informative rationale (high precision $\beta = 0.5 + \gamma$ and low recall error $\alpha = 0.5 - \gamma$, where $0 < \gamma < 0.5$)}: In this regime, it can be shown based on dereivative analysis that as $\gamma$ increases (more informative rationale), the terms involving $\gamma$ in the numerators and denominators of the conditional entropies become more prominent. The mutual information will increase with $\gamma$, as the rationale becomes more informative about the preference.

\subsection{Theorem \ref{thm:genbound}: Generalization Bound}
\label{app:pfthm1}

% \ruoxi{move below to appendix }
Next, we analyze the sample complexity of training language models with and without rationales to predict preferences. We consider two regimes: 
\emph{1)} \textbf{Training with rationale}: Let $\theta_{\text{ra}}=\mathcal{A}_{\text{ra}}(D)\sim P_{\theta_{\text{ra}}|D}$ denote the parameters of the language model trained to predict $Z$ given $S$ and $R$. \emph{2)} \textbf{Training without rationale}: Let $\theta_{\text{un}}=\mathcal{A}_{\text{un}}(D_{\setminus R})\sim P_{\theta_{\text{un}}|D_{\setminus R}}$ denote the output parameters trained to predict $Z$ given only $S$, where $D_{\setminus R}$ is a dataset $D$ with rationales removed. Given a loss function $\ell$ that measures the prediction of preference $Z$, the (mean) generalization error is $\mathrm{gen}(\mu,\mathcal{A}) = \mathbb{E}_{D,\theta\sim\mathcal{A}(D)}|\mathbb{E}_\mu[\ell(\theta)] - \mathbb{E}_D[\ell(\theta)]|$, where $\mathbb{E}_\mu[\ell(\theta)]$ is the expected loss on the true distribution (true risk) and $\mathbb{E}_D[\ell(\theta)]$ is the empirical risk.

We introduce the following conditions on the relationship between $S$, $R$, and $Z$, and the learning process: \emph{1)} $H(R | Z)\leq \eta_1$ and $H(Z | R)\leq \eta_2$, i.e., the rationale $R$ is informative about $Z$ (small $\eta_2$) without excessive irrelevance (small $\eta_1$). \emph{2)} $I(\theta_{\text{ra}}; S | Z, R) \leq \delta$ for some small positive constant $\delta$, i.e., the learned model $\theta_{\text{ra}}$ does not capture much additional information from $S$ beyond what $Z$ and $R$ already provide.  Condition 1 is supported by an effective procedure to generate useful rationale $R$. To justify Condition 2, if the learning algorithm is designed to focus on capturing the information in $R$, which is highly informative about $Z$ (per Condition 1 for small $\eta_2$), we can show that the model $\theta_{\text{ra}}$ can accurately predict $Z$ without \emph{needing} to capture much additional information from $S$ beyond what is already present in $Z$ and $R$. We provide rigorous but partial justification in Appendix~\ref{app:just}.

\begin{theorem}[Generalization bounds]
\label{thm:genbound}
    Suppose the loss function $\ell$ is $\sigma$-subgaussian under the true data distribution. Under conditions 1 and 2, we have:
\begin{align}
\mathrm{gen}(\mu,\mathcal{A}_{\text{ra}}) &\leq \sqrt{\frac{2\sigma^2}{n} \cdot (I(\theta_{\text{ra}}; Z) + \delta + \eta_1 )}, \\
\mathrm{gen}(\mu,\mathcal{A}_{\text{un}}) &\leq \sqrt{\frac{2\sigma^2}{n} \cdot (I(\theta_{\text{un}}; Z) + I(\theta_{\text{un}}; S | Z) )}.
\end{align}
% \ruoxi{missing $\delta$ in the second inequality? double check} \jin{no need for probability statement. see original paper. we define delta as: $I(\theta_{\text{ra}}; S | Z, R) \leq \delta$ for some small positive constant $\delta$, i.e., the learned model $\theta_{\text{ra}}$ does not capture much additional information from $S$ beyond what $Z$ and $R$ already provide}
\end{theorem}

The proof relies on the mutual information-based generalization bounds \citep{russo2016controlling,xu2017information} and the decomposition of the mutual information terms for both training regimes using the chain rule (see Appendix \ref{app:pfthm1}). The terms $I(\theta_{\text{ra}}; Z)$ and $I(\theta_{\text{un}}; Z)$ can be expected to be similar as long as both regimes achieve good prediction of $Z$. Under the conditions of the theorem, we can observe that the sample complexity reduction depends on the gap between $I(\theta_{\text{un}}; S | Z)$ and $\delta + \eta_1$; training with rationales can lead to improved sample efficiency when the rationale does not contain irrelevant information other than those predictive of the preference $Z$, i.e., $\eta_1$ is small, and the learning process only captures the rationale information that is useful to predict $Z$. The theoretical insights are supported by our experimental results. For instance, our evaluation in Section~\ref{sec:quality_analysis} demonstrates that a detailed rationale achieves lower sample efficiency compared to more general rationales, which contain less irrelevant information beyond what is predictive of the preference; furthermore, we showed that irrelevant rationales, i.e., a large value of $\eta_1$, indeed hamper learning.

For the proof, recall that given a loss function $\ell$, the (mean) generalization error is $\mathrm{gen}(\mu,\mathcal{A}) = \mathbb{E}_{D,\theta\sim\mathcal{A}(D)}|\mathbb{E}_\mu[\ell(\theta)] - \mathbb{E}_D[\ell(\theta)]|$, where $\mathbb{E}_\mu[\ell(\theta)]$ is the expected loss on the true distribution (true risk) and $\mathbb{E}_D[\ell(\theta)]$ is the empirical risk. For fair comparison between $\mathrm{gen}(\mu,\mathcal{A}_{\text{un})}$ and $\mathrm{gen}(\mu,\mathcal{A})_{\text{ra}}$, some technical nuances arise. The key difference from the typical setup in \cite{xu2017information} is that the true data distribution $\mu$ includes the distribution for the rationale $R$, but the training regime $\mathrm{gen}(\mu,\mathcal{A}_{\text{un}})$ does not explictly use this information. However, it may seem unclear initially whether we should include that in the generalization bound, since $R$ is indeed generated based on $Z$, corresponding to the true Markov chain: $S \to Z \to R$. 

To clarify, the Markov chain for the training without rationale is: $S \to Z \to R$, with additional arrows $Z \to \theta_{\text{un}}$ and $S \to \theta_{\text{un}}$, but no arrow from $R$ to $\theta_{\text{un}}$.

Intuitively, we should account for this difference by arguing that, conditioned on the preference $Z$, the learned model $\theta_{\text{un}}$ is conditionally independent of $R$. However, due to this difference, it seems prudent to reason from first principles.

Let's start by choosing the distributions $P$ and $Q$ for the Donsker-Varadhan variational representation of the KL divergence. We set $P = P_{S,R,Z,\theta_{\text{un}}}$ and $Q = \mu^n \otimes P_{\theta_{\text{un}}}$, where $\mu$ is the distribution for $(S,R,Z)$. Then, for any measurable function $f$, we have:
\begin{align}
D(P \| Q) \geq \mathbb{E}_P[f(S,R,Z,\theta)] - \log \mathbb{E}_{(\bar{D},\bar{R},\bar{Z},\bar{\theta}) \sim Q}[e^{f(\bar{S},\bar{R},\bar{Z},\bar{\theta})}]. \label{eq:dv}
\end{align}
Now, choose $f(S,R,Z,\theta) = \lambda(\ell_D(\theta) - \ell_\mu(\theta))$ for some $\lambda \in \mathbb{R}$, where $\ell_D(\theta)$ is the empirical loss on the dataset $D$ and $\ell_\mu(\theta)$ is the expected loss under the true distribution $\mu$. Substituting this into \eqref{eq:dv}, we get:
\begin{align}
D(P \| Q) &\geq \lambda(\mathbb{E}[\ell_D(\theta)] - \mathbb{E}[\ell_\mu(\theta)]) - \log \mathbb{E}_{(\bar{D},\bar{R},\bar{Z},\bar{\theta}) \sim Q}[e^{\lambda(\ell_{\bar{D}}(\bar{\theta}) - \ell_\mu(\bar{\theta}))}] \nonumber \\
&\geq \lambda(\mathbb{E}[\ell_D(\theta)] - \mathbb{E}[\ell_\mu(\theta)]) - \frac{\lambda^2\sigma^2}{2n} \label{eq:dv_bound},
\end{align}
where the second inequality follows from the fact that $\ell_{\bar{D}}(\bar{\theta})$ is $\sigma/\sqrt{n}$-subgaussian under $Q$, due to the subgaussian assumption on the loss function.

As \eqref{eq:dv_bound} holds for any $\lambda \in \mathbb{R}$, it must also hold for the $\lambda$ that minimizes the right-hand side. This minimum occurs at $\lambda^* = n(\mathbb{E}[\ell_D(\theta)] - \mathbb{E}[\ell_\mu(\theta)])/\sigma^2$, yielding:
\begin{align*}
D(P \| Q) &\geq \frac{n}{2\sigma^2}(\mathbb{E}[\ell_D(\theta)] - \mathbb{E}[\ell_\mu(\theta)])^2.
\end{align*}

Taking the square root of both sides, we have:
\begin{align}
\mathrm{gen}(\mu,\mathcal{A}_{\text{un}}) \leq \sqrt{\frac{2\sigma^2}{n} D(P \| Q)}. 
\nonumber
\end{align}
The key observation here is that $P_{S,R,Z,\theta_{\text{un}}} = P_{\theta_{\text{un}}|S,R,Z} P_{S,R,Z} = P_{\theta_{\text{un}}|S,Z} P_{S,R,Z}$, since $\theta_{\text{un}}$ is conditionally independent of $R$ given $S$ and $Z$ in this training regime. Therefore,
\begin{align}
D_{\text{KL}}(P \| Q) &= D_{\text{KL}}(P_{S,R,Z,\theta_{\text{un}}} \| \mu^n \otimes P_{\theta_{\text{un}}}) 
\nonumber
\\
&= D_{\text{KL}}(P_{\theta_{\text{un}}|S,Z} P_{S,Z} \| \mu_{\setminus R}^n \otimes P_{\theta_{\text{un}}}) 
\nonumber
\\
&= I(S,Z;\theta_{\text{un}}),
\nonumber
\end{align}
where $\mu_{\setminus R}$ denotes the marginal distribution of $\mu$ over $S$ and $Z$ (i.e., excluding $R$).

Hence, we have:
\begin{align*}
\mathrm{gen}(\mu,\mathcal{A}_{\text{un}}) \leq \sqrt{\frac{2\sigma^2}{n} \cdot I(S,Z;\theta_{\text{un}})}. 
\nonumber
\end{align*}
Note that we use $I(S,Z;\theta_{\text{un}})$ instead of $I(S,R,Z;\theta_{\text{un}})$, which is the key difference from the typical setup in \cite{xu2017information}, where the learned model is assumed to depend on all the data.

We can then arrive at the result for training without rationale by noting:
\begin{align}
I(\theta_{\text{un}}; S, Z) &= I(\theta_{\text{un}}; Z) + I(\theta_{\text{un}}; S | Z) . 
\nonumber
\end{align}

For training with rationale, we have:
\begin{align}
I(\theta_{\text{ra}}; S, R, Z) &= I(\theta_{\text{ra}}; Z) + I(\theta_{\text{ra}}; R | Z) + I(\theta_{\text{ra}}; S | Z, R) 
\nonumber
\\
&\leq I(\theta_{\text{ra}}; Z) + H(R | Z) + I(\theta_{\text{ra}}; S | Z, R) 
\nonumber
\\
&\leq I(\theta_{\text{ra}}; Z) + \eta_1 + \delta ,
\nonumber
\end{align}
where the first inequality is due to $I(\theta_{\text{ra}}; R | Z) \leq H(R | Z)$, and the second inequality follows from conditions 1 and 2. We can now apply [Thm. 1]\cite{xu2017information} to yield the result.

\subsection{Supporting Lemmas}
\label{app:just}
\begin{lemma}
\label{lem:fano}
     Let $\hat{Z}$ be the estimate of $Z$ based on $\theta$. Let $P_e=P(Z\neq \hat{Z}|\hat{Z},\theta)$ be the probability of error in predicting $Z$ using $\theta$. Suppose $P_e\leq \epsilon$. Then, we have that:
    \begin{equation}
        P_e\geq \frac{H(Z)-I(R;\theta)-I(Z;\theta|R) -H(\epsilon)}{\log |Z|}.
    \end{equation}
\end{lemma}
\textbf{Proof: } First, let's define an indicator variable $E$ for the error event: $E = \begin{cases} 
      0, & \text{if}\ \hat{Z} = Z \\
      1, & \text{if}\ \hat{Z} \neq Z
   \end{cases}.$ We have
\begin{align}
    H(Z|\theta) &= H(Z|\hat{Z},\theta)=H(Z,E|\hat{Z},\theta)
\nonumber
    \\
    &=H(E|\hat{Z},\theta)+H(Z|E,\hat{Z},\theta)
\nonumber
\\
    &=H(P_e)+P_eH(Z|E=1,\hat{Z},\theta)+(1-P_e)H(Z|E=0,\hat{Z},\theta)
\nonumber\\
    &\leq H(\epsilon)+P_e\log |Z|.
\nonumber
\end{align}
Hence, we have 

\begin{equation}
    \label{eq:P_e}
    P_e\geq \frac{H(Z|\theta) -H(\epsilon)}{\log |Z|}=\frac{H(Z)-I(Z;\theta) -H(\epsilon)}{\log |Z|}.
\end{equation}
This part of the  proof follows from Fano's inequality. 

Now, since $I(Z;\theta)+I(R;\theta|Z)=I(Z,R;\theta)$, we have 
\begin{align}
    I(Z;\theta)&=I(Z,R;\theta)-I(R;\theta|Z)
    \nonumber
    \\
    &=I(R;\theta)+I(Z;\theta|R)-I(R;\theta|Z)
    \nonumber
    \\
    &\leq I(R;\theta)+I(Z;\theta|R)
    \nonumber
\end{align}
Plugging in \eqref{eq:P_e} obtains the result.

Note: $I(Z;\theta|R)=0$ if $\theta$ is trained only on $R$, i.e., $Z$ and $\theta$ are conditionally independent given $R$. 

\begin{lemma}
\label{lem:upper}
    Let $\hat{Z}$ be the estimate of $Z$ based on $\theta$. Let $P_e=P(Z\neq \hat{Z}|\hat{Z},\theta)$ be the probability of error in predicting $Z$ using $\theta$. Assume that $P_e\leq 0.5$. Then, we have that:
    \begin{equation}
        P_e\leq \frac{H(Z)-I(R;\theta)+H(R|Z)}{\log 2}.
    \end{equation}
\end{lemma}

\textbf{Proof:} By definition of $E$ from Lemma \ref{lem:fano}, we have 
\begin{align}
    H(Z|\theta) &= H(Z|\hat{Z},\theta)=H(Z,E|\hat{Z},\theta)
    \nonumber
    \\
    &=H(E|\hat{Z},\theta)+H(Z|E,\hat{Z},\theta)
    \nonumber
    \\
    &\geq H(E|\hat{Z},\theta)
    \nonumber
    \\
    &=P_e\log(1/P_e) + (1-P_e)\log(1/(1-P_e))
    \nonumber
    \\
    &\geq P_e\log 2
    \nonumber
\end{align}
Hence, we have $P_e\leq H(Z|\theta)/\log 2$. Since $H(Z|\theta)= H(Z) - I(Z;\theta)$, and that  
\begin{align}
    I(Z;\theta)&=I(Z,R;\theta)-I(R;\theta|Z)
    \nonumber
    \\
    &=I(R;\theta)+I(Z;\theta|R)-I(R;\theta|Z)
    \nonumber
    \\
    &\geq I(R;\theta)-H(R|Z)+H(R|\theta,Z)
    \nonumber
    \\
    &\geq I(R;\theta)-H(R|Z),
    \nonumber
\end{align}
the bound follows.

\textbf{Partial justification of Condition 2}: Consider the Markov chain $Z \rightarrow R \rightarrow \theta$ with additional arrows of $Z \rightarrow \theta$ and $R \rightarrow \theta$, where $Z$ is the preference, $R$ is the rationale, and $\theta$ is a trained model used to predict $Z$. From both Lemma \ref{lem:fano} and \ref{lem:upper}, we see that as long as $\theta$ captures the information of $R$, i.e., $I(R;\theta)$ is large, and $R$ does not contain excessive irrelevant information other than $Z$, i.e., $H(R|Z)$ is small, the prediction error of $Z$ from model $\theta$ can be well-controlled. Specifically, based on the lower and upper bound analysis, we can conclude that the probability of error $P_e$ decreases with increasing $I(R;\theta)$ or decreasing $H(R|Z)$. This implies that the model does not need to capture additional information from $S$ to achieve high prediction accuracy for $Z$, i.e., $I(\theta_{\text{ra}}; S | Z, R)$ is small. In other words, the incorporation of $R$ in the training of $\theta_{\text{ra}}$ guides the model to easily predict $Z$ without resorting to finding potentially irrelevant information from $S$. A full justification of the condition hinges on a detailed analysis of the specific algorithm and is beyond the scope of this study.

\subsection{Proof of Theorem \ref{thm:mutual_info} and Derivation of Regimes}

Recall the definition of $\alpha, \beta, \epsilon, f(\cdot)$ in Sec. \ref{app:MIanalysis}. Now, we derive the relationship between the mutual information $I(Z; R | S)$ and these parameters:
%\jin{use $z$ instead of $k$ below for consistency.}
\begin{align}
    P(Z=1 | S,R=1) &= \frac{P(R=1|Z=1,S)P(Z=1|S)}{P(R=1|S)} 
    \nonumber
    \\
    &=\frac{P(R=1|Z=1,S)P(Z=1|S)}{\sum_{z\in \{0,1\}}P(R=1|Z=z,S)P(Z=z|S)} 
    \nonumber
    \\    &=\frac{\beta(f(S)+\epsilon)}{\beta (f(S) + \epsilon) + \alpha (1 - f(S) - \epsilon)}.
    \nonumber
\end{align}
Similarly, we have 
\begin{align}
    P(Z=1 | S,R=0) &=\frac{\alpha(f(S)+\epsilon)}{\alpha (f(S) + \epsilon) + \beta (1 - f(S) - \epsilon)}.
    \nonumber 
\end{align}
These equations show that the probability of $Z = 1$ given the query $X$, the preferred response $Y_1$, the dispreferred response $Y_2$,
%\jin{provide definitions. Also note you used $Y1$ and $Y2$ in A.1. It's better to remain consistency.}, 
and the rationale $R$ depends on both the informativeness of the rationale, through $\alpha$ and $\beta$, and the informativeness of the query and responses, through $f(S)$. Using the above equations, we get the conditional entropies as follows:
\begin{align}
    H(Z| S,R=1) = H\left(\frac{\beta(f(S)+\epsilon)}{\beta (f(S) + \epsilon) + \alpha (1 - f(S) - \epsilon)} \right),
    \nonumber
    \\
    H(Z| S,R=0) = H\left(\frac{\alpha(f(S)+\epsilon)}{\alpha (f(S) + \epsilon) + \beta (1 - f(S) - \epsilon)} \right)
    \nonumber
\end{align}
and substituting to the mutual information equation, we get:
\begin{align}
    I(Z;R|S) = H(Z|S) - \sum_{r= \{0,1\}} P(R=r|S)H(Z|S,R=r).
    \nonumber
\end{align}
Then, we compute each probabilities $P(R|S)$ as follows:
%\jin{use $S$ below instead of $X,Y_w, Y_l$}
\begin{align}
    P(R = 1 | S) &= \sum_{z \in \{0,1\}} P(R=1 |Z=z, S)P(Z=z|S) 
    \nonumber
    \\
    &= \beta (f (S)+\epsilon) + \alpha (1 - f(S) - \epsilon), 
    \nonumber
    \\
    P(R = 0 | S) &= 1 - P(R = 1 | S) 
    \nonumber
    \\
    &= 1 - (\beta (f (S)+\epsilon) + \alpha (1 - f(S) - \epsilon))
    \nonumber
\end{align}
and substitute them back into the conditional mutual information term, where we define $p = f(S)$:
% \jin{define $p$. Also check throughout the paper to remove the equation numbers unless the equation has been referred to. This can make the numbering more thoughtful and declutter the writing. Only number the key equations in your paper to make them stand out.}
\begin{align}
    I(Z;R | S) = H(p+\epsilon)
    -&(\beta (p+\epsilon) + \alpha (1 - p-\epsilon)) H\left(\frac{\beta (p+\epsilon)}{ \beta (p+\epsilon) + \alpha (1 - p-\epsilon)}\right) 
    \nonumber
    \\
    &- (1 - (\beta (p+\epsilon) + \alpha (1 - p-
\epsilon))) H\left(\frac{\alpha (p+\epsilon)}{ \alpha (p+\epsilon) + \beta (1 - p-\epsilon)}\right).
    \nonumber
\end{align}

To study the influence of the parameters $\alpha, \beta, \epsilon$ on the mutual information, we consider the following edge cases:

\paragraph{Regime 1: Highly informative rationale $\beta \approx 1$ and low noise $\alpha \approx 0$ $\implies$ Rationale is a sufficient statistics.} \mbox{}\\
In this case, the conditional probabilities are simplified to
\begin{align}
    P(Z = 1 | S, R = 1) \approx \frac{f(S) + \epsilon}{f(S) + \epsilon} = 1, 
    \nonumber
    \\
    P(Z = 1 | S, R = 0) \approx \frac{0}{1-f(S) - \epsilon} = 0.
    \nonumber
\end{align}

Thus, the mutual information becomes as follows:
%\jin{use $S$}
\begin{align}
    I(Z; R | S) \approx H(f(S) + \epsilon) - P(R = 1 | S) H(1) 
    \nonumber
    \\
    - P(R = 0 | S) H(0)  = H(f(S) + \epsilon).
    \nonumber
\end{align}

In this regime, the conditional mutual information is solely determined by the entropy of the preference prediction, $H(f(S) + \epsilon)$. We notice that the entropy function is concave and reaches the max value at the $0.5$ mark.

\paragraph{Regime 2: Uninformative rationale  $\beta \approx 0.5$ and high noise $\alpha \approx 0.5$.} \mbox{}\\

For this case, the conditional probabilities become:
\begin{align}
    P(Z = 1 | S, R = 1) \approx \frac{f(S) + \epsilon}{f(S) + \epsilon + 1 - f(S) -\epsilon} = f(S) + \epsilon  = P(Z = 1 | S, R = 0)
    \nonumber
\end{align}
and the mutual information equals to:
\begin{align}
    I(Z; R | S) \approx H(f(S) + \epsilon) - H(f(S) + \epsilon) = 0,
    \nonumber
\end{align}
which shows that the rationales provides 0 information about the preference given the prompt and responses.

\paragraph{Regime 3: Moderately informative rationale  $\beta = 0.5 + \gamma $ and lower noise $\alpha = 0.5 - \gamma$.} \mbox{}\\

Given the assumption of $\beta = 0.5 + \gamma$ and $\alpha = 0.5 - \gamma$, where $0 \leq \gamma \leq 0.5$ and $\gamma$ denotes the level of informativeness of the rationale, we substitute this into the conditional mutual information term.

We first substitute into the conditional probabilities and get:
\begin{align}
    P(Z = 1 | S, R = 1) = \frac{(0.5 + \gamma) (f(S) + \epsilon)}{(0.5 + \gamma) (f(S) + \epsilon) + (0.5 - \gamma) (1 - f(S) - \epsilon)},
    \nonumber
    \\
    P(Z = 1 | S, R = 0) = \frac{(0.5 - \gamma) (f(S) + \epsilon)}{(0.5 - \gamma) (f(S) + \epsilon) + (0.5 + \gamma) (1 - f(S) - \epsilon)}.
    \nonumber
\end{align}
Then, we compute the following probabilities:
\begin{align}
    P(R = 1 | S) &= \sum_{z \in \{0,1\}} P(R=1|Z=z,S)P(Z=z|S) 
    \nonumber
    \\
    &= 0.5 + 2\gamma (f(S) + \epsilon -0.5),
    \nonumber
    \\
    P(R = 0 | S) &= 0.5 - 2\gamma (f(S) + \epsilon -0.5).
    \nonumber
\end{align}

Now, we can compute the conditional mutual information term:
%\jin{Please draw a graph or a series of graphes to show the value below as a function of $p=f(S) + \epsilon$ and $\gamma$. The graph should be enought o support your reasoning of regimes and trends below.}
\begin{align}
    I(Z;& R | S) = H(f(S) + \epsilon) 
    \nonumber
    \\
    & -(0.5 + 2\gamma (f(S) + \epsilon - 0.5))\cdot H \left(  \frac{(0.5 + \gamma) (f(S) + \epsilon)}{(0.5 + \gamma) (f(S) + \epsilon) + (0.5 - \gamma) (1 - f(S) - \epsilon)}\right)
    \nonumber
    \\
    & -(0.5 - 2\gamma (f(S) + \epsilon - 0.5))\cdot H \left(  \frac{(0.5 - \gamma) (f(S) + \epsilon)}{(0.5 - \gamma) (f(S) + \epsilon) + (0.5 + \gamma) (1 - f(S) - \epsilon)}\right).
    \label{eq: mutual info}
\end{align}

\begin{figure}[h]
  \begin{center}
    \includegraphics[width=0.8\textwidth]{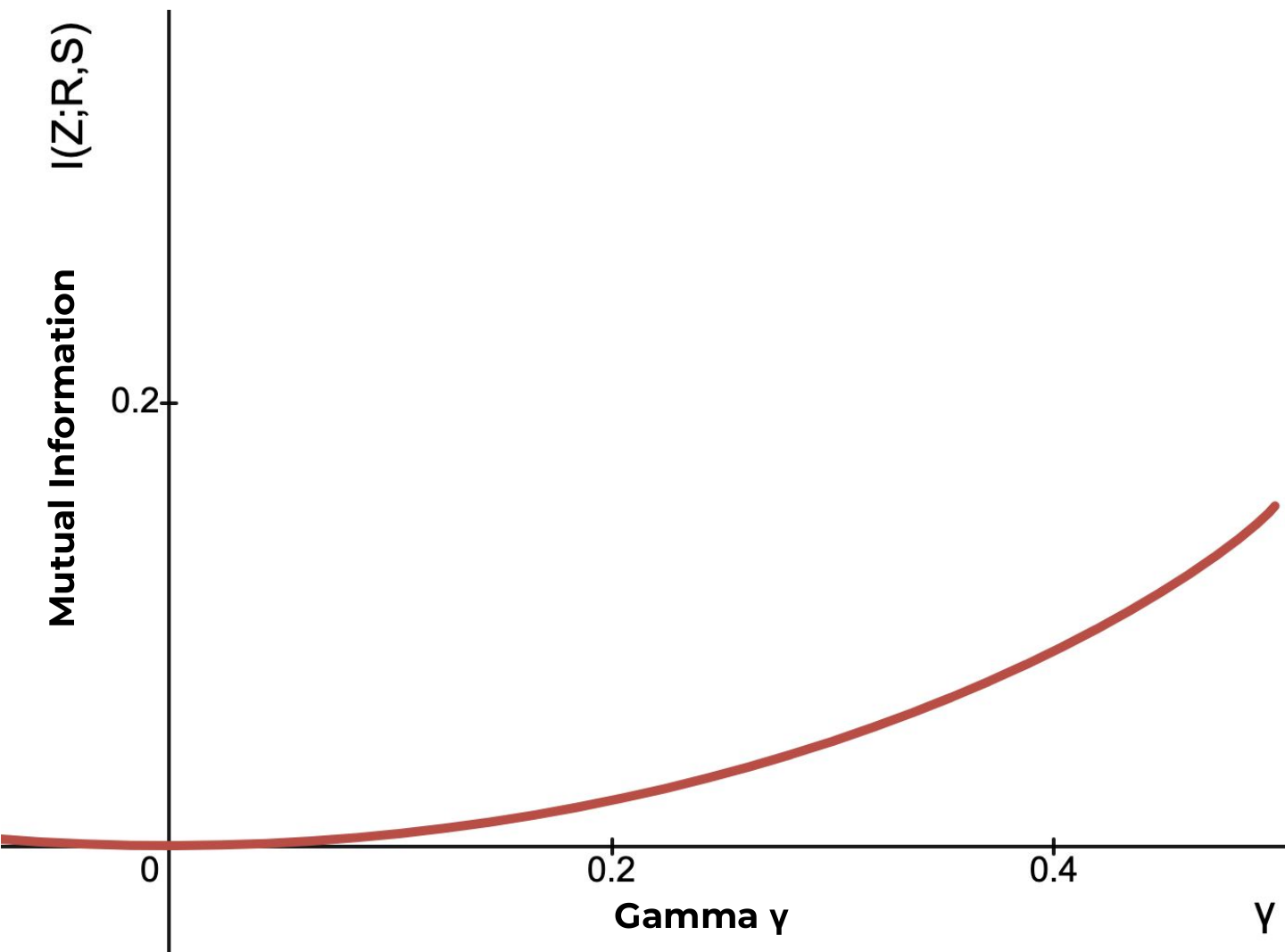}
  \end{center}
  \caption{The plot of Equation~\ref{eq: mutual info} showing the relation between mutual information and gamma $\gamma$ for a fixed $f(S)$.
}
  % \ruoxi{shall we move it to intro}}
    \label{fig:mutual info}
\end{figure}

\begin{figure}[h]
  \begin{center}
    \includegraphics[width=0.8\textwidth]{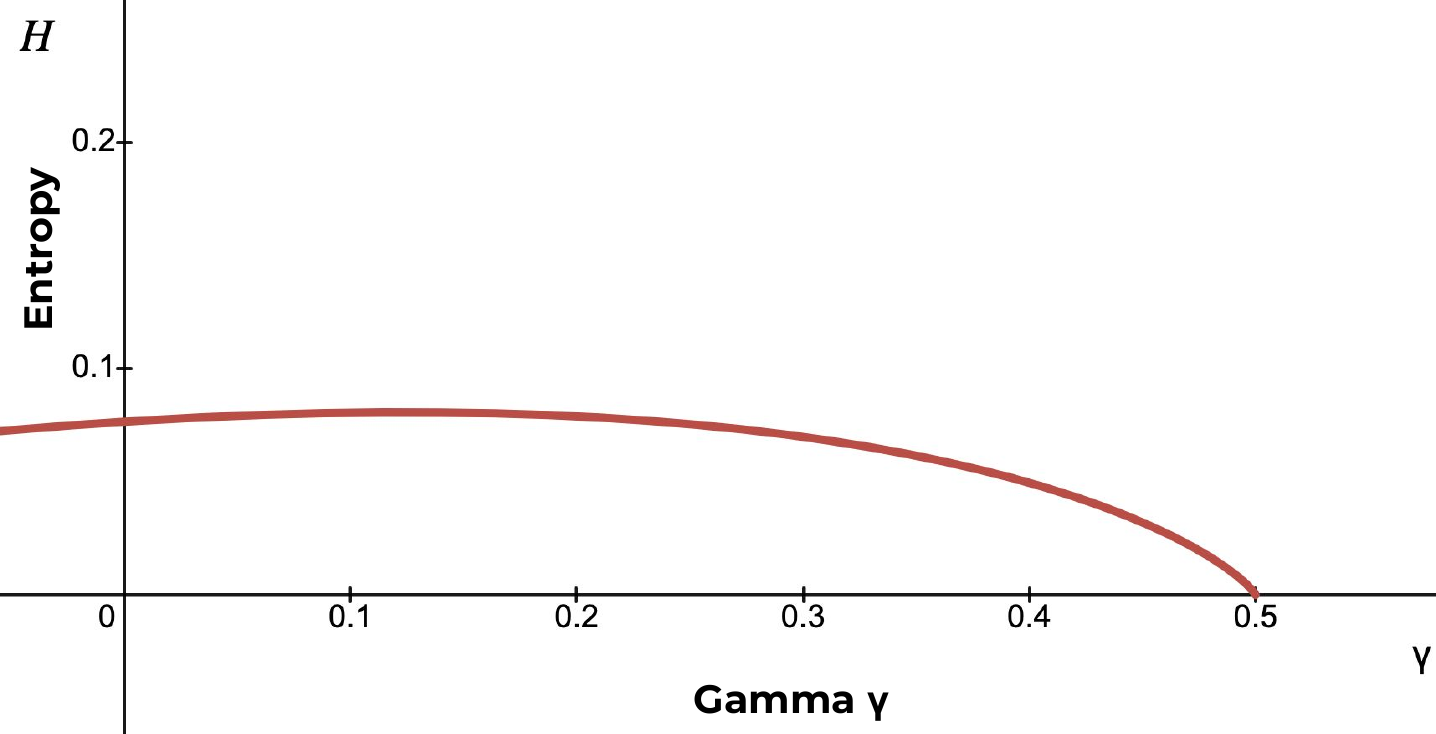}
  \end{center}
  \caption{The plot of Equation~\ref{eq: mutual info} showing the relation between the first entropy term and gamma $\gamma$ for a fixed $f(S)$.
}
  % \ruoxi{shall we move it to intro}}
    \label{fig:entrop 1}
\end{figure}

\begin{figure}[h]
  \begin{center}
    \includegraphics[width=0.8\textwidth]{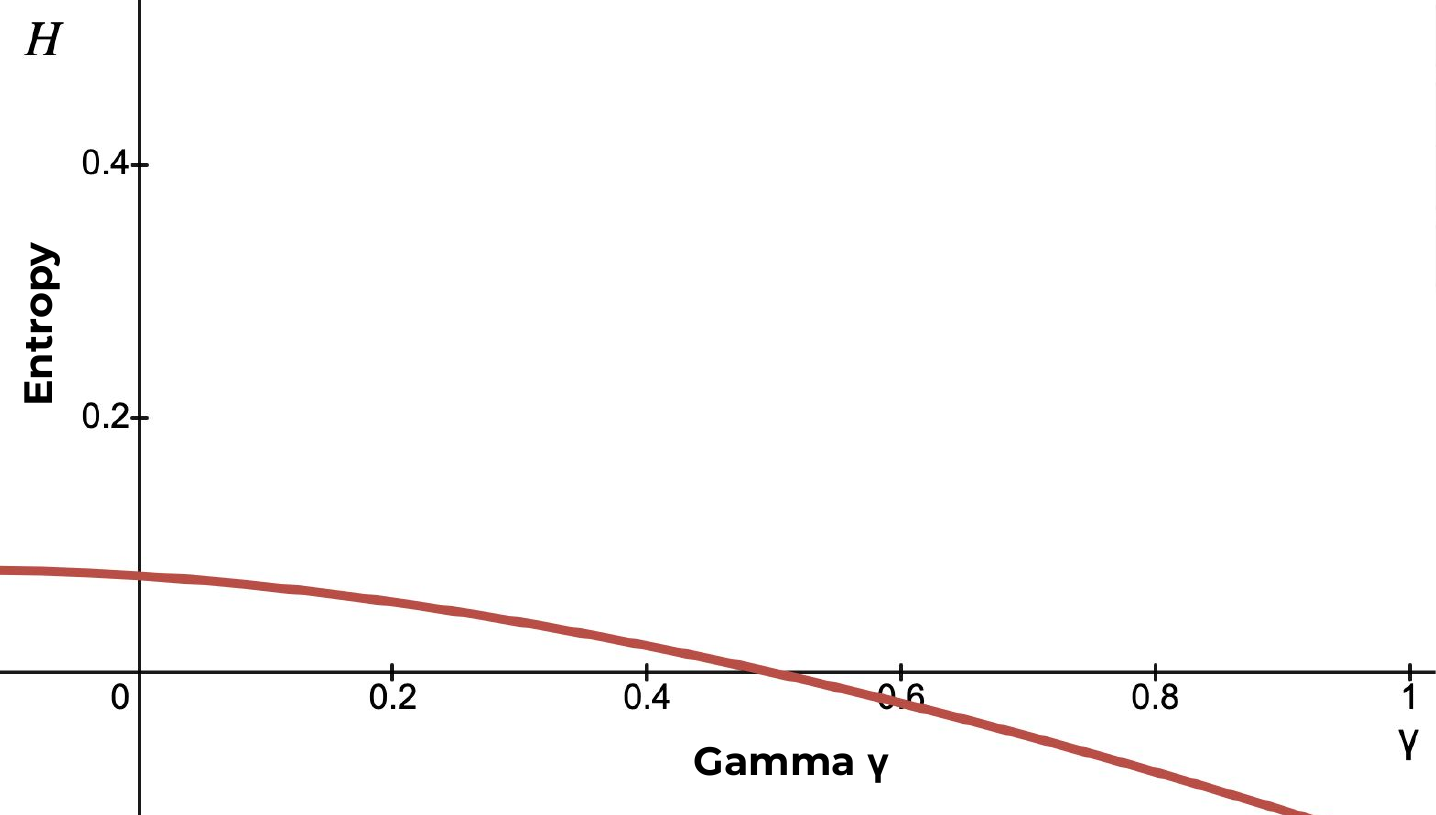}
  \end{center}
  \caption{The plot of Equation~\ref{eq: mutual info} showing the relation between second entropy term and gamma $\gamma$ for a fixed $f(S)$.
}
  % \ruoxi{shall we move it to intro}}
    \label{fig:entrop 2}
\end{figure}

We can now analyze the behavior of the mutual information as a function of $\gamma$:

\textbf{When the rationale is uninformative $\gamma = 0$}, then the mutual information becomes $0$, $I(Z;R | S) = 0$, which is consistent with previous cases, in which uninformative rationales provide no additional information about the preference $Z$ as demonstrated in Figure~\ref{fig:mutual info}. 

{As rationale becomes more informative about the preference by increasing $\gamma$, we observe that mutual information also increases displayed in Figure~\ref{fig:mutual info}.}

Consider the case that the true probability $P(Z = 1 | S) = f(S) + \epsilon > 0.5$, so the preference $Z$ is most likely to be $1$. 

Then, we now focus on terms in $I(Z;R | S)$ that contain $\gamma$. As $\gamma$ increases, the first entropy weight term $(0.5 + 2\gamma (f(S) + \epsilon - 0.5)) $ increases and the second entropy weight term  $(0.5 - 2\gamma (f(S) + \epsilon - 0.5)) $ decreases. Entropy terms also involve $\gamma$, and we observe that as $\gamma$ increase, the ratio $\frac{(0.5 + \gamma) (f(S) + \epsilon)}{(0.5 + \gamma) (f(S) + \epsilon) + (0.5 - \gamma) (1 - f(S) - \epsilon)}$ approaches $1$, since the numerator grows faster than the denominator. Thus, the first entropy term decreases with $\gamma$, as the entropy of a distribution to a deterministic one is lower.

Conversely, for the second entropy term, with the increase of $\gamma$, the ratio $\frac{(0.5 - \gamma) (f(S) + \epsilon)}{(0.5 - \gamma) (f(S) + \epsilon) + (0.5 + \gamma) (1 - f(S) - \epsilon)}$ approaches $0$, so the second entropy term also decreases with $\gamma$.

\begin{itemize}
    \item For the first entropy term, with an increase of $\gamma$, the weight of the first entropy term increases, but the entropy decreases itself (see Figure~\ref{fig:entrop 1}). 
    \item For the second entropy term, with an increase of $\gamma$, the weight of the second entropy term decreases, and the entropy decreases itself (see Figure~\ref{fig:entrop 2}). 
\end{itemize}

The net effect on the mutual information depends on the relative magnitudes of these changes. However, we can argue that the decrease in the entropy terms dominates the change in their weights due to the entropy function changing more rapidly near the extremes (i.e., when the distribution is close to being deterministic) compared to the middle range. 

Thus, with an increase in $\gamma$, the overall contribution of the entropy terms to the mutual information decreases, causing an increase in $I(Z; R | S)$, which indicates that as the rationale becomes more informative about the preference, the mutual information increases.

\newpage

\section{Additional Experimental Results}

\subsection{Experimental Details} \label{app: exp details}

For DPO-based methods, we fine-tune the base model by supervised fine-tuning (SFT) with the chosen responses from the preference dataset for a single epoch. For ORPO, which avoids the reference model, we skip this SFT step. For models trained on RDPO and results reported in Section~\ref{sec:exp}, we use $\gamma = 2.0$, and for RORPO, we use $\gamma = 10.0$. Similar to baseline methods, we train RDPO and RORPO for $1$ epoch. We perform ablation studies on the hyperparameter $\gamma$ and the number of epochs in the following sections.

For our winrate scores, we report the mean winrates after querying the evaluator $3$ times. To reduce the evaluator's order bias, we have additionally shuffled the order of responses. We note that the winrate error bars are within $<3\%$ for $512$ samples.

\subsection{Ablation Study on Models and Hyperparameters}\label{app:models and hyperparameters}

\begin{table}[h!]
\centering
\begin{tabular}{ccccccc}
      
         & \multicolumn{6}{r}{Mistral-7B-v0.1    }      \\
         \cmidrule[0.9pt](l{0.525em}r{0.525em}){2-7}
General  & 62                       & 61  & 63  & 61  & 62  & 60   \\
Detailed & 64                       & 66  & 63  & 61  & 60  & 62   \\
\midrule
$\gamma$   & 1.0                      & 1.5 & 2.0 & 2.5 & 3.0 & 10.0 \\
\\
         & \multicolumn{6}{r}{Mistral-7B-v0.2-Instruct}    \\
         
         \cmidrule[0.9pt](l{0.525em}r{0.525em}){2-7}
General  & 55                       & 62  & 57  & 55  & 57  & 56   \\
Detailed & 56                       & 56  & 57  & 60  & 57  & 59   \\
\midrule
$\gamma$   & 1.0                      & 1.5 & 2.0 & 2.5 & 3.0 & 10.0 \\
\\
         & \multicolumn{6}{r}{Zephyr-7B-Beta}   \\

         \cmidrule[0.9pt](l{0.525em}r{0.525em}){2-7}
General  & 58                       & 55  & 55  & 57  & 55  & 57   \\
Detailed & 48                       & 49  & 51  & 51  & 50  & 51   \\
\midrule
$\gamma$   & 1.0                      & 1.5 & 2.0 & 2.5 & 3.0 & 10.0 
      
\end{tabular}
\caption{The impact of different values of hyperparameter $\gamma$ on the winrate of the RDPO model against the DPO model. The results on various models: Mistral-7B-v0.1 \textbf{(Top)}, Mistral-7B-Instruct-v0.2 \textbf{(Middle)}, and Zephyr-7B-Beta \textbf{(Bottom)}.}
\label{tab:more_gamma}
\end{table}

Here, we investigate the impact of the hyperparameter $\gamma$, ranging from $1.0$ to $10.0$, on the performance of the model trained with RDPO loss. We provide the winrate scores against the DPO model on the Orca dataset. As we see in Table~\ref{tab:more_gamma}, models trained on either general or detailed rationales can still achieve a stronger winrate against the DPO model, except for the case of \emph{Zephyr-7B-Beta} model, which achieves a draw with the DPO model. For this model, the better quality rationales are important for effective preference learning, as the general rationales can still improve the performance.

\subsection{Ablation on the Number of Epochs}\label{app:more epochs}

\begin{table}[h!]
\centering
\begin{tabular}{lcc}
        & General & Detailed \\
        \cmidrule[0.9pt](l{0.525em}r{0.525em}){2-2}
        \cmidrule[0.9pt](l{0.525em}r{0.525em}){3-3}
Epoch 1 & 76      & 74       \\
Epoch 2 & 75      & 71       \\
Epoch 3 & 74      & 74      
\end{tabular}
\caption{The analysis of the number of epochs on the performance. Winrate of the RDPO model trained on the general rationales \textbf{(left)} and detailed rationales \textbf{(right)} against the DPO model, respectively.}
\label{tab:more_epochs}
\end{table}

In the main paper, we trained the models with the RDPO loss on a single epoch similar to DPO~\citep{rafailov2024direct}. Here, we study the impact of training the models with the rationales on more epochs. As we observe in Table~\ref{tab:more_epochs}, training with more epochs does not improve the winrate of the RDPO model against the DPO model. The reason could be that the model has already learned the preference well after the first epoch, or it could be that the quality of rationales could be further improved to increase the efficiency of the rationale-based preference learning algorithms.

\subsection{Rationale Quality Analysis}
\label{sec:quality_analysis}

In this section, we examine the importance of the quality of rationales for preference learning. We study different types of rationales and possible errors encountered in rationales, and how these affect the preference learning of the model.

% General vs Specific Rationales
% \begin{figure*}
% \begin{wrapfigure}{r}{0.70\textwidth}
\begin{figure}[h]
% \vspace{-2.5em}
    \centering%
	\includegraphics[width=1.00\linewidth]{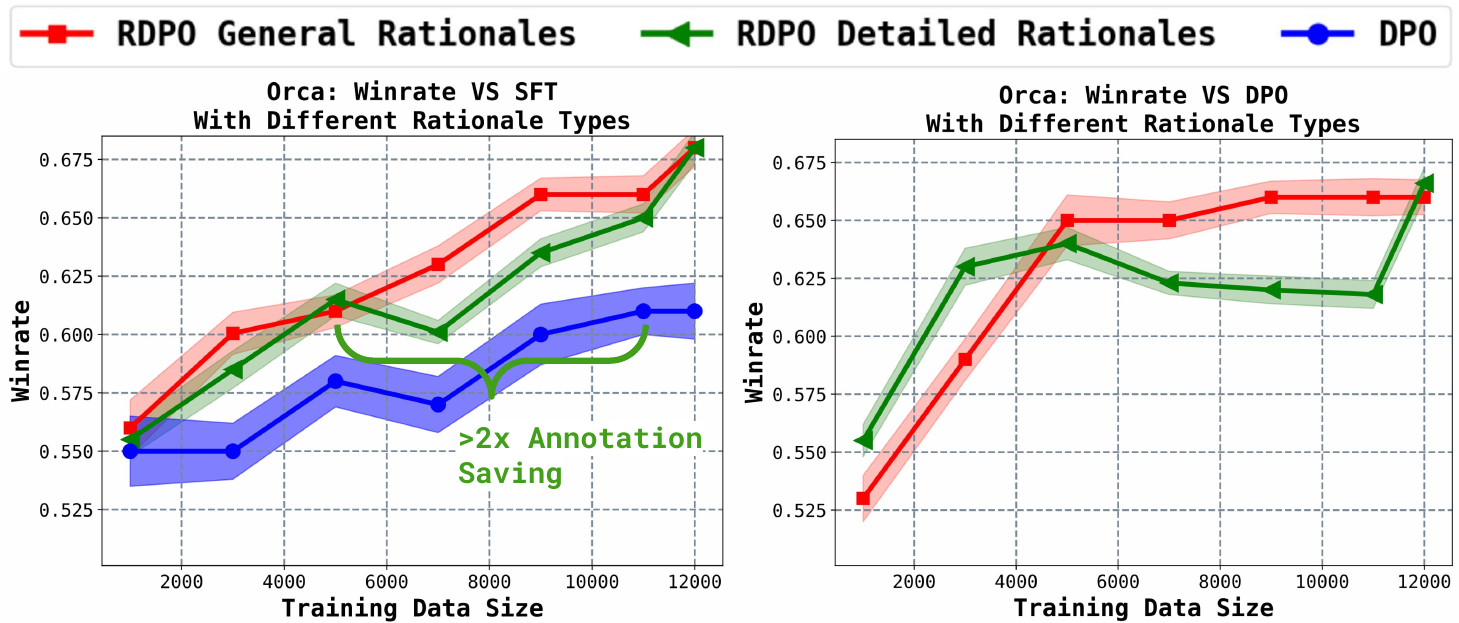}
	\caption{Measuring the impact of types of rationales on the RDPO performance. \textbf{Left:} Comparing the winrates against the SFT model. \textbf{Right:} Comparing the winrates against the DPO model.} 
 % \ruoxi{illustrate data saving for the first plot}}
    
    \label{fig:detailed rationales}
% \end{figure*}
\end{figure}

\paragraph{Detailed Rationales vs General Rationales.} Explaining why one answer is preferred over the other can be expressed in multiple ways through many perspectives. Here, we study the level of granularity of the rationales, general (which explains the preference at a high level without going into details) and detailed (which explains in details and pinpoints specifically to the prompt and the response). 
% \ruoxi{query}).
We use language models to automatically generate the rationales for the Orca dataset according to our intent. For details on generation prompts, we refer to Appendix~\ref{app:rationale generation}. We provide samples of these rationales in Appendix~\ref{app:rationale comparison}. After training the models on respective rationales, we compare the winrates between RDPO trained models and the DPO one. Figure~\ref{fig:detailed rationales} on the left shows that the model trained on general rationales with the RDPO loss converges earlier to a high winning rate against the SFT model than the model trained on the detailed rationales. The reason could be that the general rationales share common features across the samples (e.g., clarity, conciseness, directness), which lets the model learn quickly and transfer these learning cues to other samples more easily, while detailed rationales might require more time to fully comprehend them. However, in both cases, the models trained on these rationales reach better winrates than the DPO against the SFT model. In Figure~\ref{fig:detailed rationales} (right), both RDPO models can have a better winrate $>57\%$ against the DPO model with as few as $3{,}000$ training samples, while the DPO model is trained on $11{,}000$ samples. 
% \ruoxi{need to cmopare with the samples required by DPO, or using percentage to show the data efficiency gain more directly} 
We provide results on models trained on additional epochs in Appendix~\ref{app:more epochs}.

\begin{table}[]
    \centering
\resizebox{0.95\linewidth}{!}{
    \begin{tabular}{cc|cc}
       \multicolumn{2}{c}{Permuted Rationales VS Original Rationales} & \multicolumn{2}{c}{Opposite Rationales VS Original Rationales} \\
       \cmidrule[0.9pt](l{0.525em}r{0.525em}){1-2}
       \cmidrule[0.9pt](l{0.525em}r{0.525em}){3-4}
        \multicolumn{2}{c}{1 \quad \quad \quad \quad\quad \quad \quad \quad 99} & \multicolumn{2}{c}{10 \quad \quad \quad \quad\quad \quad \quad \quad 87}  
    \end{tabular}
    }
    \vspace{1em}
    \caption{The analysis of quality of rationales on the RDPO performance. The winrate comparison between the RDPO models trained on rationales with errors and original rationales. \textbf{Left:} Permuted, irrelevant rationales. 
    \textbf{Right:} Opposite, inaccurate rationales.}
    \label{tab:rationale quality}
\end{table}

% \paragraph{Impact of rationale quality}

% We show the importance of the quality of rationales 

\paragraph{Low-Quality Rationales.} RDPO has shown efficacy with rationales generated by the off-the-shelf models, even when the models have not undergone preference alignment.
% high-quality, correct rationales. \ruoxi{we don't know whether the rationales we used are correct, tone down this part like RDPO shows effecicy with rationales generated by off-the-shelf models, even when the models have not undergone preference alignment} 
However, we want to further analyze the impact of rationale's quality on RDPO's performance.
% \ruoxi{the impact of rationale's quality on RDPO's performance.}whether RDPO would still perform with low/bad-quality rationales. 
In particular, we examine the rationale quality in terms of its relevance and correctness. One case of a low-quality rationale can be a completely irrelevant rationale to the given pair of responses. 
% For this case, 
To simulate irrelevant rationales,
% \ruoxi{to simulate irrelevant rationales} 
we permute the abovementioned detailed rationales over different samples so that no rationale is relevant to the context. Training the model on these rationales with RDPO and comparing one trained on original rationales, we show in Table~\ref{tab:rationale quality} that it achieves less than $1\%$ winrate against the RDPO model trained on correct and relevant rationales. 
% For another case of low-quality rationales, 
% \ruoxi{To study the impact of correctness,} 
To study the impact of correctness, we negate the general rationales to have the opposite meaning and observe that the RDPO model trained on original rationales gets almost a $90\%$ winrate. As we note, the quality of rationales is important for improving the preference learning performance. 
% \ruoxi{While we showed that the rationales generated by off-the-self LMs can already bring significant improvement to preference learning, we expect that more deliberate control of the rationale quality can further improve preference learning. We leave the in-depth exporation of strategies for quality-controlled rationale generation to future work . }
While we showed that rationales generated by the off-the-shelf language models can already bring significant improvement to preference learning, we expect that more deliberate control of the rationale quality can further improve preference learning. We leave the in-depth exploration of strategies for generating quality-controlled rationales to future work.
% Therefore, apart from collecting high-quality pairwise preferences, high-quality rationales can further improve preference learning.

% quality of rationales

% hard to control quality of rationales so we design

% irrelevant
% accuracy of

% We show that with more epoch, the performance might improve but not as much show in appendix.

% Rationales source ablation

% Permute Rationales and Make opposite rationales from the original one, we show winrates

% \begin{wrapfigure}{r}{0.45\textwidth}
% \begin{figure}[h]
% % \vspace{-2.5em}
%     \centering%
% 	\includegraphics[width=0.52\linewidth]{fig/ablation-model-squeeze.png}
% 	\caption{Studying the source of rationale generation (\textbf{Y-axis}) for the Orca dataset. Using rationales from different sources to train the models on RDPO (\textbf{X-axis}). Winrate of the RDPO against the DPO model. 
%     }
%     \label{fig:ablation source}

% \end{figure}

\begin{table}[h!]
\centering
\resizebox{0.65\linewidth}{!}{
\begin{tabular}{lcc}
      
    Generated By     & \multicolumn{2}{c}{RDPO vs DPO Winrate} \\
   \cmidrule[0.9pt](l{0.525em}r{0.525em}){1-1}
         \cmidrule[0.9pt](l{0.525em}r{0.525em}){2-3}
Mistral-7B-Instruct-v0.2     & 76-23 & 50-46 \\
Llama3-8B-Instruct     & 73-27 & 52-45 \\
Phi3-Mini-4K    & 75-25 & 51-49 \\
\cmidrule[0.9pt](l{0.525em}r{0.525em}){2-3}
     & Mistral-7B-Instruct-v0.2 & Llama3-8B-Instruct \\
% \midrule
   & \multicolumn{2}{c}{Trained On} \\

\end{tabular}
}
\caption{Studying the impact of different source of rationale generation (\textbf{Y-axis}) for the Orca dataset on the model training with RDPO (\textbf{X-axis}). Winrate of the RDPO model against the DPO model. }
\label{tab:ablation source}
\end{table}

\paragraph{Rationale Source.} While collecting the human-annotated rationales could be high-quality, in practice, it is costly with time and resources. Therefore, we resolve to language models to generate rationales. In our experiments, we use the base models to create them. Here, we study the rationales coming from other sources and how they impact the RDPO training. We generate rationales for the Orca dataset on three different models: Mistral-7B-Instruct-v0.2, LLama3-8B-Instruct, and Phi3-Mini-4K~\cite{abdin2024phi}. Then, we use these rationales to train the first two models. Results from Figure~\ref{tab:ablation source} show us consistent winrates against the DPO model with slightly better winrate from the same source as the base model. This shows that the rationales can be transferred to other models for preference training with rationales. Especially, leveraging models with small sizes $\{3,7,8\}$ billion parameters, we can generate rationales to improve preference learning. \update{However, we observe modest improvements on the Llama-3.1-8B-Instruct experiment, which can be attributed to two key factors. First, the inherent capability of Llama-3.1-8B-Instruct surpasses Mistral-7B-Instruct-v0.2, making substantial gains more challenging to achieve on this stronger baseline. Second, the general preference datasets like Orca and Ultrafeedback, which include pre-existing responses, may not be fully optimized for Llama-3.1-8B-Instruct.}
% Further studies on high-quality rationales are of high importance. 

% - source of rationale (different models)

\paragraph{Quality of Generated Rationales.}
Here, we provide the quality of generated rationales judged by the GPT-4o in terms of helpfulness, relatedness, correctness, and coherence. As we observe in Table~\ref{table: eval ratio}, the rationales generated are of good quality. We also shuffled the rationales to random prompts and also changed the meaning of rationales for sanity check of our evaluator and we observe significant drops in the scores. Please see \ref{app:evaluate rationale prompt} for prompting details.

\begin{table}[h!]
\centering
\resizebox{0.85\linewidth}{!}{
\begin{tabular}{lllll}
                           & Helpfulness & Relatedness & Correctness & Coherence \\
                           \toprule
Generated Rationales       & 4.99        & 4.99        & 4.99        & 5.00      \\
Shuffled Rationales        & 3.22        & 1.98        & 2.20        & 4.2       \\
Changed Meaning Rationales & 2.14        & 2.63        & 1.58        & 3.2      
\end{tabular}
}
\caption{The quality scores for generated rationales from the LLaMA-3.1-8B-Instruct model for UltraFeedback dataset. The scores are in the range from 1-5. }
\label{table: eval ratio}
\end{table}

% Not need the same LLM, aligning different LLM with other LLMs can still work
% transferable 

% \paragraph{Impact of specificity of rational (Genernal vs specific) and why general works better}

% Two hypotheses for why
% - model capability is not enough to understand specific rationales
% - not enough compute to digest specific rationales

% \paragraph{ablation study}

\update{\subsection{Response Comparison}}

% \begin{wraptable}{r}{7.0cm}
% \vspace{-1em}
\begin{table}[h]
\centering
\resizebox{0.55\linewidth}{!}{
\begin{tabular}{lcccc}
     & \multicolumn{2}{c}{Avg Output Length}               & \multicolumn{2}{c}{TriviaQA (Exact Match)} \\ 
     \cmidrule[0.9pt](l{0.525em}r{0.525em}){2-3}
     \cmidrule[0.9pt](l{0.525em}r{0.525em}){4-5} 
      & DPO & RDPO & DPO & RDPO \\
      \toprule 
      Orca & 2021 & 1364 & 34.9 & 35.7 \\
      UltraFeedback & 2066 & 1299 & 31.5 & 33.1 \\
      % Llama3-8B Instruct - Orca & 2 & 3 & 4 & 5 
      \bottomrule
\end{tabular}
}
% \vspace{1em}
    \caption{Comparison between DPO and RDPO. \textbf{Left:} The average output lengths of the generated responses on the prompts from the test sets of respective datasets. \textbf{Right:} The exact match (EM) performance on the TriviaQA dataset of the preference-trained models on respective datasets.}
\label{table: triviaqa}
% \vspace{-1em}
\end{table}
% \end{wraptable}

% \paragraph{Response Comparison.} 
We compare the responses generated by the DPO- and RDPO-trained models. As shown in Table~\ref{table: triviaqa}, the average output length by the DPO trained model is much longer than the RDPO trained model in the case of the Orca dataset, which is more than $60\%$ longer on average. Due to longer output, there might be a chance for a higher occurrence of hallucinations. Therefore, we want to study the correctness of the outputs from these models. For this reason, we use the TriviaQA~\citep{joshi2017triviaqa} dataset and using LM Evaluation Harness~\citep{eval-harness} to measure the exact match (EM) accuracy and compare between the models. We see in Table~\ref{table: triviaqa} that models trained with the DPO loss experience a decrease in performance, compared to the models with RDPO loss. As a reason, we emphasize the importance of measuring the hallucinations in the generations of both models in future studies. We provide a comparison of the responses from the two models in Appendix~\ref{app: dpo vs rdpo responses} and a time cost analysis in Appendix~\ref{app: cost and time} to help model owners better understand the trade-offs of our method.

\update{\subsection{Cost Analysis}}
\label{app: cost and time}

In this section, we analyze the cost breakdown to assist project owners in evaluating trade-offs. Specifically, we present a cost-benefit analysis of the approach. The table below outlines the cost of using the API to generate rationales for a given number of annotations. It also highlights the RDPO win rate compared to the SFT model for each data budget and estimates the number of annotations potentially saved by using RDPO to achieve the same level of performance as DPO.

\begin{table}[]
    \centering
    \begin{tabular}{l|r|r|r|r|r}
         API Rationale Cost &	\$0.13 	&  \$0.19 	& \$0.26 	& \$0.32 	& \$0.39 \\
         Annotations Used 	& 1K 	& 1.5K 	& 2K 	& 2.5K & 	3K \\
         Annotations Saved 	& 3K 	& 6K 	& 6.5K 	& 6.8K 	& >10K \\
vs SFT Winrate 	& 54\% 	& 56\% 	& 58\% 	& 60\% 	& 62\%
    \end{tabular}
    \caption{Cost-analysis breakdown of using RDPO with the OpenAI API (gpt-4o-mini model) to generate rationales, highlighting performance improvements relative to the number of annotations used and annotations saved.}
    \label{tab:cost analysis}
\end{table}

While open-weight models were used to generate the rationales in our study, the table illustrates the associated costs of utilizing an API model, specifically gpt-4o-mini. The pricing for this model is \$0.150 per 1M input tokens and \$0.600 per 1M output tokens. The results shown in Table~\ref{tab:cost analysis} are based on the Mistral-7B-v0.2-Instruct model trained on the Orca dataset.

We also report the runtime for RDPO and DPO for one epoch on the Llama-3.1-8B-Instruct model using 12,000 Orca examples, as follows:

\begin{itemize}
    \item RDPO General: 6770 seconds
    \item RDPO Specific: 6950 seconds
    \item DPO: 3583 seconds.
\end{itemize}

While processing additional tokens nearly doubles the runtime, RDPO compensates for this by requiring fewer annotations while achieving comparable or superior performance to DPO. Furthermore, we report the average response lengths for the Orca dataset, highlighting that rationales are approximately 50\% shorter in length compared to chosen or rejected responses:

\begin{itemize}
    \item Chosen responses: 786
    \item Rejected responses: 981
    \item Rationale responses: 411
\end{itemize}

\subsection{Rationale Generation} \label{app:rationale generation}

% Show Prompts to generate rationales
Here, we provide the prompts to generate detailed and general rationales as mentioned in Section~\ref{sec:exp}. In Prompt~\ref{prompt:general}, we ask the model to generate general rationales without going into details and provide high-level reasoning. In Prompt~\ref{prompt:specific}, we ask the model to provide a specific rationale and go as much into detail as possible. These are our examples of prompts and we call for further studies to improve the prompts for higher quality rationales.

\begin{center}
\fbox{
\begin{minipage}{38em}
  \texttt{Given the prompt:\\
  <prompt>\\
  Which of the following assistant's responses is preferred and strictly follows the prompt question? Why this assistant's response: \\
  <chosen>\\
  is preferred over the response provided below:\\
  <rejected>\\
  On a high and general level, why the response above is preferred over the response below? Provide a general, high-level explanation for your reasoning without going into the response's details.}
\end{minipage}
} 
\captionof{Prompt}{Prompt to the LLM to retrieve a general rationale.\label{prompt:general}}

\end{center}

\begin{center}
\fbox{
\begin{minipage}{38em}
  \texttt{Given the prompt:\\
  \textit{<prompt>}\\
  Which of the following assistant's responses is preferred and strictly follows the prompt question? Why this assistant's response: \\
  \textit{<chosen>}\\
  is preferred over the response provided below:\\
  \textit{<rejected>}\\
  Why the response above is preferred over the response below? Provide a specific, detailed explanation for your reasoning pointing directly at the response.}
\end{minipage}
} 
\captionof{Prompt}{Prompt to the LLM to retrieve a detailed rationale.\label{prompt:specific}}

\end{center}
\subsection{Comparison of Rationales } \label{app:rationale comparison}

We provide examples of the general and detailed rationales for the Orca dataset by the Mistral-7B-Instruct-v0.2, and compare them in Examples~\ref{example:0},~\ref{example:1}, and~\ref{example:2}. As we observe, the general and detailed rationales follow our intent. Additionally, we observe that a part of the general rationale is also included in the detailed rationale. It will be important to study, which type of rationale is more helpful for the preference training.

\begin{center}

\fbox{
\begin{minipage}{38em}
  \texttt{\textit{<General Rationale>}\\
  The first response strictly adheres to the prompt by providing the RDF triplets in the required format, while the second response includes an explanation that goes beyond the required format and may not be necessary for the task at hand. Thus, the first response is preferred as it focuses solely on providing the required output in the correct format. \\
  \\
  \textit{<Detailed Rationale>}\\
  The assistant's response is preferred because it follows the standard format for RDF triplets, where the subject comes first, followed by the predicate, and then the object. In the given input sentence, 'AFC Ajax (amateurs)' and 'Ajax Youth Academy' are the subjects, 'has ground' and 'plays at' are the predicates, and 'Sportpark De Toekomst' is the object that is common to both subjects. Therefore, the RDF triplets should be structured as [subject, predicate, object], with each triplet representing a distinct relationship. The assistant's response correctly reflects this structure and semantics, making it a preferred option.}
\end{minipage}
} 
\captionof{Example}{An example $\#1$ of general and detailed rationales.\label{example:0}}
\end{center}

\begin{center}
\fbox{
\begin{minipage}{38em}
  \texttt{\textit{<General Rationale>}\\
  The preferred response strictly follows the prompt question by calculating the total amount paid per episode for both major and minor characters based on the given information. It uses the provided information to calculate the salaries of the major and minor characters and then adds them together to find the total amount paid per episode. The response below includes unnecessary details and makes the response longer and less efficient. \\
  \\
  \textit{<Detailed Rationale>}\\
  The assistant's response above is clear and accurate. They first calculate the total amount paid to the minor characters per episode, which is \$60,000. Then, they calculate the total amount paid to the major characters per episode, which is \$225,000. Finally, they add the salaries of the major and minor characters to find the total amount James pays per episode, which is \$285,000."
  This explanation directly points to the steps taken in the assistant's response and explains each step in detail. It also uses clear and concise language to make the explanation easy to understand. }
\end{minipage}
} 
\captionof{Example}{An example $\#2$ of general and detailed rationales.\label{example:1}}
\end{center}
\begin{center}

\fbox{
\begin{minipage}{38em}
  \texttt{\textit{<General Rationale>}\\
  The response above is preferred because it directly answers the question by specifically identifying the kidneys as the organs that secrete hormones to help maintain homeostasis, while the response below goes into excessive detail and provides additional information that is not directly related to the question. \\
  \\
  \textit{<Detailed Rationale>}\\
  The assistant's response is preferred because it provides a clear and detailed explanation of the hormones secreted by the kidneys and their roles in maintaining homeostasis. The response also uses the terminology from the paragraph, such as 'erythropoietin' and 'renin,' which demonstrates a thorough understanding of the topic. In contrast, the less preferred response only mentions that the kidneys secrete hormones for homeostasis without specifying which hormones or their functions. Therefore, the preferred response is more informative and accurate, making it a better answer for the given question. }
\end{minipage}
} 
\captionof{Example}{An example $\#3$ of general and detailed rationales.\label{example:2}}
\end{center}

winrate using HALOs repository \cite{ethayarajh2023halos}

\subsection{Evaluating Responses with LLM as a Judge}\label{app:evaluate winrate}

\begin{center}

\fbox{
\begin{minipage}{38em}
  \texttt{Here is a conversation between \textit{<user>} and  \textit{<assistant>}.\\
  \textit{<history>}\\
  \textit{<assistant>} can now choose from different responses.  \\
  Response 1: \textit{<response1>}\\
  END OF RESPONSE 1\\
  Response 2: \textit{<response2>}\\
  END OF RESPONSE 2\\
  Between these two responses, which response is more helpful, harmless, and concise? Pick one.}
\end{minipage}
} 
\captionof{Prompt}{Prompt to the LLM (e.g., GPT-4o) to retrieve the preference over two responses.\label{prompt:llm preference}}

\end{center}
After training the models with preference learning algorithms, we want to compare the performance between the models. Thus, a common practice is to generate the responses from respective models and compare the quality between them. As a human evaluator might be expensive to acquire, we usually use a capable language model to judge the responses. In our case, we leverage the GPT-4o model to judge and provide winrate scores. We provide the prompt to determine a preferred response between generations in Prompt~\ref{prompt:llm preference}, which is implemented in the HALOs code implementation~\citep{ethayarajh2023halos} and shuffles responses to avoid the ordering bias and averaged the scores over the runs.

\subsection{Evaluating Generated Rationale with LLM as a Judge}\label{app:evaluate rationale prompt}

\begin{center}

\fbox{
\begin{minipage}{38em}
\texttt{
Evaluate the quality of the rationale comparing the preference pairs for a given prompt. 
  \\
Provide the score through three categories of helpfulness, correctness, and coherence with a range from 1 to 5, where:\newline
\\
Helpfulness:\newline
5: Rationale provides significant insights and valuable guidance.\newline
4: Rationale is useful and relevant, contributing positively to understanding.\newline
3: Rationale offers some relevant information, but limited usefulness.\newline
2: Rationale provides minimal assistance, possibly tangential.\newline
1: Rationale is unhelpful, irrelevant, or misleading.\newline
\\
Relatedness:\newline
5: Rationale fully relates and correctly answers the initial question.\newline
4: Rationale relates correctly to the initial question with minor discrepancies.\newline
3: Rationale relates correctly to the initial question with many discrepancies.\newline
2: Rationale relates to the initial question but is tangential.\newline
1: Rationale does not relate to the initial question.\newline
\\
Correctness:\newline
5: Rationale is factually accurate and logically sound.\newline
4: Rationale is accurate with minor inaccuracies.\newline
3: Rationale contains a mix of correct and incorrect information.\newline
2: Rationale contains significant inaccuracies or flawed logic.\newline
1: Rationale is entirely false or fundamentally flawed.\newline
\\
Coherence:\newline
5: Rationale is clear, well-organized, and easy to follow.\newline
4: Rationale is generally clear and organized, with minor lapses.\newline
3: Rationale is understandable, but lacks clarity in parts.\newline
2: Rationale is difficult to follow due to poor organization.\newline
1: Rationale is incomprehensible and disorganized.\newline
\\
Prompt:\newline
$<$beginning of prompt$>$\newline
$\%$s\newline
$<$end of prompt$>$\newline
\\
Rationale:\newline
$<$beginning of rationale$>$\newline
$\%$s\newline
$<$end of rationale$>$\newline
\\
Present the scores in a JSON dictionary format: scores = $\{$"helpfulness": $\#$, "relatedness": $\#$, "correctness": $\#$, "coherence": $\#$$\}$.
}
\end{minipage}
}

\captionof{Prompt}{Prompt to the LLM (e.g., GPT-4o) to retrieve the score of a rationale.\label{prompt:llm rationale}}

\end{center}

Before training, we evaluate the generated rationales with a more capable model to check the quality of the generation. Please see the Prompt~\ref{prompt:llm rationale} for details. After verification, we drop the samples with low-scored rationales (any metric with a score 3 or less). Then, the new resulting filtered dataset, we use for further training.

\subsection{DPO and RDPO Generated Responses}
\label{app: dpo vs rdpo responses}

We provide some examples and compare responses generated by the models trained with DPO and RPDO losses, respectively. In Example~\ref{example: 1dpo vs rdpo}, we observe the case when DPO demonstrates the verbosity and generates a lengthy output compared to the response by the RDPO model. The length bias in DPO is also demonstrated in Example~\ref{example: 3dpo vs rdpo}, which generates not only a longer output than RDPO but also outputs a personal opinion on a political stance. Example~\ref{example: 2dpo vs rdpo} asks for generating a tweet for a single sentiment. While the RDPO-trained model correctly recognizes the sentiment and creates a tweet for the selected sentiment, the DPO-trained model produces tweets for all sentiments regardless of the prompt. Despite the responses generated by RDPO being shorter on average than the ones by DPO, when prompted to generate an article, the RDPO response can be as lengthy as the DPO one as shown in Example~\ref{example: 4dpo vs rdpo}.

\begin{center}

\fbox{
\begin{minipage}{38em}
  \texttt{\textit{<Prompt>}\\
  What is your opinion on Britain exiting the EU?\\
What is your opinion on Britain leaving the EU and Brexit?\\
Would you say that these questions are the same?
\\
  \\
  \textit{<DPO Response}>\\
  These questions are related but not exactly the same. Both ask for an opinion on Britain's decision to leave the European Union (EU), which is known as Brexit. However, the first question asks specifically about Britain exiting the EU, while the second question expands to include the broader concept of Brexit, which includes not only leaving the EU, but also the negotiations and potential outcomes of that decision. \\
\\
Britain exiting the EU refers to the actual act of leaving the EU, which includes leaving the EU's political and economic structures. Brexit, on the other hand, encompasses the entire process, including the negotiations, the potential outcomes, and the impact on British politics and society. \\
\\
So, if someone is asking for your opinion on Britain exiting the EU, they are asking for your thoughts on the actual act of leaving the EU. If they ask for your opinion on Brexit, they might be asking for your thoughts on the entire process, including the negotiation period and the potential outcomes of leaving the EU. \\
\\
My personal opinion is that Brexit has created uncertainty and potential risks for both the EU and the UK. However, it also gives the UK the opportunity to set its own trade and regulatory policies, which could potentially lead to new opportunities and economic growth.\\
  \\
  \textit{<RDPO Response}>\\
  Yes, I would consider the questions to be similar, but not exactly the same. The first question asks for your opinion on Britain's departure from the EU, while the second question specifically refers to Brexit, which is the name given to the process of Britain's departure from the EU. So, while the second question is a subset of the first, it adds a layer of specificity by referring to the name given to the process.
  }
\end{minipage}
} 
\captionof{Example}{A comparison of generated responses from DPO and RDPO models.\label{example: 3dpo vs rdpo}}

\end{center}

\begin{center}

\fbox{
\begin{minipage}{44em}
  \texttt{\textit{<Prompt>}\\
  I ' ve already decided that I want to live in Finland for a semester . What then ? What if - I know , I ' m getting ahead of my own dream - world - he and I are together for so long ... we fall in love ? What then ?\\
Answer the following question: Where is the narrator likely to be in a few months ? \\
  \\
  \textit{<DPO Response>}\\
  In a few months, the narrator is likely to be living in Finland for a semester as they have already decided on that.
\\
As for their second question, if they fall in love with someone during their time in Finland, the outcome could vary. The narrator could choose to continue their relationship despite the distance if they are committed to making it work. They could also choose to pursue the relationship further if both parties are interested.
\\
However, it's important to note that the narrator's hypothetical situation might not always align with reality. If they do fall in love and pursue a long-distance relationship, it could require more effort, patience, and communication to maintain the relationship. Ultimately, it depends on the individuals involved and their level of commitment to making it work.
\\
In any case, their current plan is to go to Finland for a semester, so that is where they are likely to be in a few months. 
\\
Regarding the world reference in the second question, it's not clear how it relates to the overall context of the scenario. If it refers to a fictional world, it's likely that the question is asking about the potential outcomes of an imaginary situation. 
\\
However, if it refers to the real world, it's worth noting that long-distance relationships can be challenging, but they are also possible with dedication, trust, and open communication. If the narrator decides to pursue a relationship with someone, they can consider using technology to stay connected, such as video calls, messaging, and email. They could also set aside time for regular updates on each other's lives, and plan visits to see each other as often as possible.\\
Ultimately, the key to maintaining a successful long-distance relationship is to approach it with a positive attitude, patience, and an understanding that challenges will arise. By staying committed to each other and working together to overcome any obstacles, the narrator and their partner could potentially build a strong and lasting connection. So, assuming the narrator's plan to go to Finland for a semester remains unchanged, they are likely to be in Finland in a few months. Whether or not they fall in love, the outcome depends on the individuals involved and their willingness to make the relationship work despite the challenges. 
\\
\[\text{Response continues...}\]
  \\
  \textit{<RDPO Response>}\\
  The narrator is likely to be living in Finland for a semester. However, if they fall in love with someone, their plans might change depending on their feelings and circumstances. }
\end{minipage}
} 
\captionof{Example}{A comparison of generated responses from DPO and RDPO models.\label{example: 1dpo vs rdpo}}

\end{center}

\begin{center}

\fbox{
\begin{minipage}{38em}
  \texttt{\textit{<Prompt>}\\
  Sentiment possibilities. Choices:\\
(A). negative;\\
(B). positive;. Generate a tweet that has the following sentiment: (A).
\\
  \\
  \textit{<DPO Response}>\\
  (A). negative: "I just received a faulty product from the company, and it's really frustrating! \#CustomerServiceFail"\\
  (B). positive: "Had an amazing time at the concert last night! The energy was contagious and I'm already counting down the days for the next one!\\
  \\
  \textit{<RDPO Response}>\\
  Here's a tweet that conveys a negative sentiment:\\
  "Feeling down today. The rain outside matches my mood.
  }
\end{minipage}
} 
\captionof{Example}{A comparison of generated responses from DPO and RDPO models.\label{example: 2dpo vs rdpo}}

\end{center}
\begin{center}

\fbox{
\begin{minipage}{44em}
  \texttt{\textit{<Prompt>}\\
  Write an article based on this summary:
\\
Jack Tector and Nathan Smith have received their first Ireland call-up for this month's ICC Intercontinental Cup game against the Netherlands.
\\
  \\
  \textit{<DPO Response}>\\
  Title: Fresh Blood in the Irish Squad: Jack Tector and Nathan Smith Await Their Debut against the Netherlands\\
\\
Ireland, a long-standing cricket force in European cricket, is set to feature two of its exciting young talents, Jack Tector and Nathan Smith, in this month's ICC Intercontinental Cup game against the Netherlands. These young guns will be making their debut on the international stage, providing them with an unforgettable experience and an opportunity to shine amongst the cricketing world's elite.\\
\\
Jack Tector, a 23-year-old all-rounder, has been a consistent performer in both domestic and provincial cricket, earning him a well-deserved Ireland call-up. With his skills in both batting and bowling, Tector offers immense potential to the side and can add valuable depth to the Irish squad. His aggressive approach at the crease, coupled with his ability to generate pace and bounce from the ball, make him a formidable prospect to look out for.\\
\\
Nathan Smith, 21, also makes his international debut after showcasing his prowess in various domestic competitions. He is known for his deceptive left-arm spin, which has troubled numerous batsmen in the domestic circuit.\\
  \\
  \textit{<RDPO Response}>\\
  Ireland Calls Up Jack Tector and Nathan Smith for Upcoming ICC Intercontinental Cup Game against Netherlands\\
\\
The Irish cricket team has made some exciting announcements recently, as two young and promising players, Jack Tector and Nathan Smith, have been granted their first call-ups to the national squad for the ICC Intercontinental Cup game against the Netherlands. This upcoming match is set to take place this month, and both Tector and Smith are eager to make the most of this incredible opportunity.\\
\\
Jack Tector, a 21-year-old all-rounder from Wiltshire, England, has been on the radar of Irish selectors for some time now. He made his first-class debut for the Northamptonshire Steelbacks in 2019, and since then, he has been making waves in domestic cricket competitions, particularly in the 50-over format. Tector has been consistently impressive with the bat, having scored a century and several half-centuries in various domestic games. He is also a handy left-arm spinner, which adds depth to Ireland's bowling attack.\\
\\
Nathan Smith, on the other hand, is a 23-year-old left-handed opening batsman from County Antrim, Northern Ireland.
  }
\end{minipage}
} 
\captionof{Example}{A comparison of generated responses from DPO and RDPO models.\label{example: 4dpo vs rdpo}}

\end{center}

\end{appendices}

\end{document}